\documentclass{article} % For LaTeX2e
\usepackage{iclr2027_conference,times}
\usepackage{graphicx}
\usepackage{booktabs}
\usepackage{multirow}
\iclrfinalcopy

\renewcommand{\headrulewidth}{0pt}
\usepackage{amsmath,amsfonts,bm}

\def\eqref#1{equation~\ref{#1}}
\def\1{\bm{1}}

\DeclareMathAlphabet{\mathsfit}{\encodingdefault}{\sfdefault}{m}{sl}
\SetMathAlphabet{\mathsfit}{bold}{\encodingdefault}{\sfdefault}{bx}{n}

\usepackage{hyperref}
\usepackage{url}
\usepackage{caption}

\title{Hybrid Particle-Voxel Dynamics \\ for Multimodal 3D World Simulation}

\author{Jimmy Wang$^{1,2}$ \quad
Yifan Chou$^{3}$ \quad
Kaifeng Zhang$^{1}$ \quad
Yixuan Wang$^{4}$ \\
\bf Binghao Huang$^{1}$ \quad
Tao Yu$^{5}$ \quad
Saleh Nabi$^{5}$ \quad
Yunzhu Li$^{1}$ \\[4pt]
\normalfont
$^{1}$Columbia University \quad
$^{2}$UC Berkeley \quad
$^{3}$University of Michigan \quad
$^{4}$NVIDIA \quad
$^{5}$Analog Devices
}

\begin{document}

\maketitle

\begin{figure}[h]
% \vspace{-10pt}
\includegraphics[width=\textwidth]{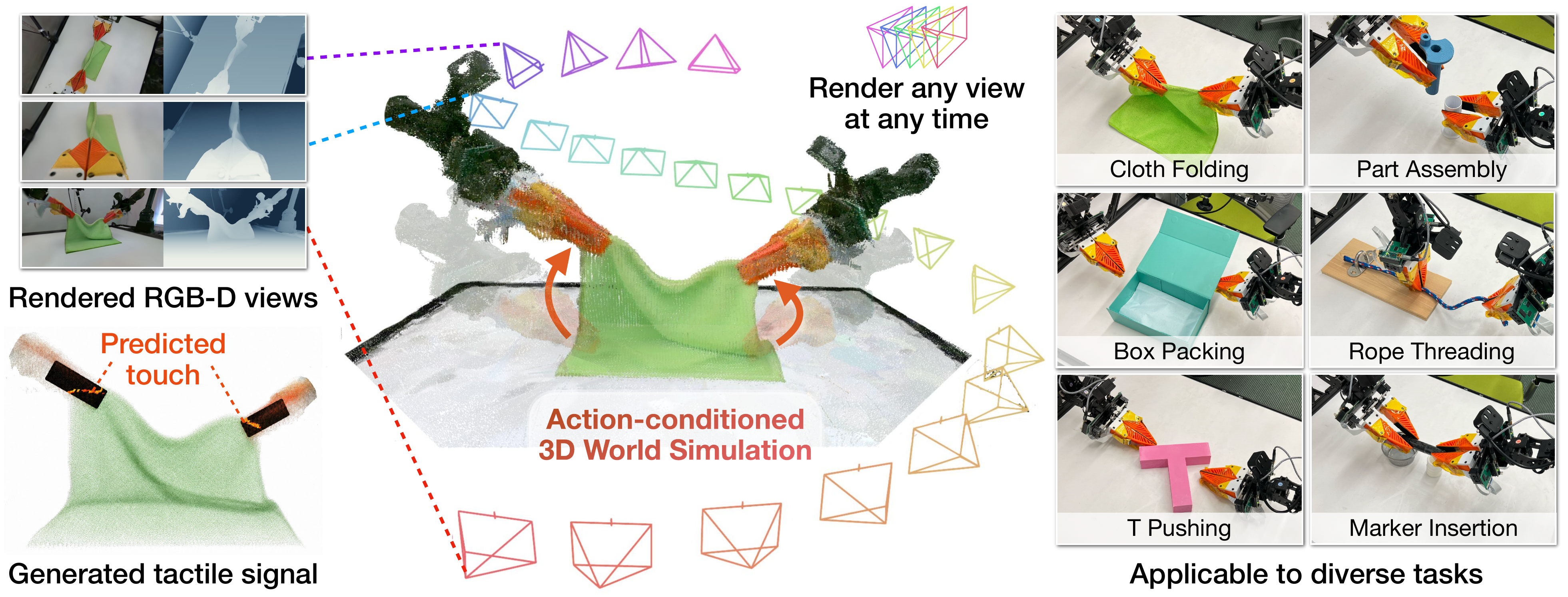}
\vspace{-15pt}
\caption{\small \textbf{Overview of 3D World Simulator.} We introduce an action-conditioned, multimodal 3D world modeling framework that operates in a unified 3D representation space, jointly predicting 3D geometry (middle), multi-view consistent RGB and depth from novel views (upper left), and spatially aligned tactile signals (bottom left). 3D World Simulator is applicable to a diverse set of tasks, supporting interactive, long horizon data generation at 50 Hz for policy training and evaluation. Project Page: \url{https://jwang771.github.io/3dworldsimulator/}.}
\end{figure}

\begin{abstract}
Predicting how the world evolves in response to robot actions is central to policy learning, evaluation, and planning. However, many action-conditioned world models predict future observations in 2D image or video space without explicitly modeling the underlying 3D scene, making spatial consistency across viewpoints and over long-horizon interactions challenging. We introduce 3D World Simulator, an action-conditioned multimodal world modeling framework that jointly predicts appearance, geometry, and tactile observations from a unified 3D representation. Our approach combines a hybrid particle-voxel autoencoder that compresses point clouds into compact 3D voxel latents with an action-conditioned latent dynamics model that predicts their evolution. The shared 3D representation supports geometrically consistent rendering from arbitrary viewpoints and flexible integration of spatially aligned tactile signals. Our experiments demonstrate interactive, long-horizon, multimodal prediction at 50 Hz. We then show how our model enables scalable data generation for policy training and evaluation using tactile and multi-view visual observations, including wrist-mounted and novel camera views. Furthermore, we scale our method to large datasets containing diverse objects and demonstrate model-based planning with unseen objects and robot configurations. Together, these results highlight the potential of hybrid particle-voxel dynamics as a foundation for multimodal 3D world simulation.

\end{abstract}

\section{Introduction} 

Predicting how the 3D world evolves in response to interaction is a fundamental capability for intelligent robotic systems, with applications to data generation for policy training~\citep{wang2026interactive}, policy evaluation~\citep{guo2026ctrl}, model-based planning~\citep{huang2026pointworld}, and as backbones for world-action modeling~\citep{nvidia2025cosmos}. Existing action-conditioned world models are predominantly built using 2D video generation backbones. While able to produce photorealistic predictions, these models do not explicitly represent the underlying 3D scene, making it difficult to maintain spatial consistency across viewpoints and to model the precise spatial relationship between contact and geometry. Prior work has explored using 3D representations~\citep{huang2026pointworld, zhang2024dynamic}, but these methods typically rely on strong structural assumptions, such as object-specific state representations or explicit correspondences, that can be unreliable in complex interactions and limit broader applicability.

In this work, we present \textbf{3D World Simulator}, an action-conditioned multimodal world modeling framework that predicts in a \textbf{unified 3D representation space}, providing a common interface capable of generating 3D particles, spatially aligned tactile signals, and multi-view consistent renderings from novel viewpoints. Our model supports interactive, long horizon, multimodal 3D video prediction at 50 Hz, and generalizes to diverse objects and environments when trained at scale. 

3D World Simulator consists of two technical components. \textbf{(1) Hybrid particle-voxel autoencoder.} The encoder takes point clouds obtained directly from RGB-D video and compresses them into a compact 3D voxel latent. The decoder progressively upsamples this latent into higher-resolution voxels and then dense 3D particles in a coarse-to-fine manner. Each particle parameterizes a 3D Gaussian that can be rasterized from arbitrary viewpoints~\citep{kerbl2023gaussian}, while tactile signals are represented directly as 3D particles within the same output space~\citep{huang20243d,huang2025vt}. \textbf{(2) 3D latent dynamics model.} We learn an action-conditioned dynamics model that predicts the next 3D voxel latent directly. Because this latent encodes the scene as a whole, the dynamics model does not rely on object-specific state representations or explicit correspondences.

A key advantage of our inherently 3D world model is that a single predicted scene state creates a common interface for multiple concurrent capabilities. Since appearance is grounded in a shared 3D Gaussian representation, our model can \textbf{(1) produce geometrically and multi-view consistent renderings}, including from novel and dynamic wrist-mounted camera viewpoints. The same spatial representation provides a \textbf{(2) flexible and principled way of incorporating multimodal signals}. Tactile predictions can be represented directly as 3D particles, remaining geometrically aligned with contacted surfaces. The shared 3D representation also makes our predictions \textbf{(3) composable with existing 3D assets}, including various static backgrounds, objects, and kinematic robot models, enabling additional controllability over the simulated observations from our model. 

The 3D World Simulator enables several important downstream applications. We demonstrate \textbf{interactive, long-horizon, multimodal 3D video prediction} across a diverse set of six tasks involving rigid, articulated, and deformable objects, outperforming state of the art 2D video and 3D baselines. Next, 3D World Simulator enables \textbf{policy training and evaluation}, generating synthetic data to train imitation learning policies that rely on visuo-tactile observations, as well as multi-view observations including novel and wrist-mounted camera views. Our model can further be used to evaluate policies, demonstrating strong correlation with real evaluations. Furthermore, 3D World Simulator is \textbf{scalable to diverse objects}. We train 3D World Simulator on the large-scale Deform360~\citep{li2026deform360} dataset, a large-scale multi-view visuo-tactile dataset. Our method outperforms state-of-the-art 3D baselines and can be used for zero-shot model-based planning on unseen objects and robot arms. 

In summary, our main contributions are: (1) 3D World Simulator, a general action-conditioned multimodal world modeling framework for interactive, long horizon 3D video prediction, (2) hybrid particle-voxel dynamics in a unified 3D representation space, enabling view consistency and flexible integration of multimodal signals, and (3) a unified model that enables a range of applications, including policy training, policy evaluation, and planning. 

\section{Related Works}

\paragraph{2D Action-Conditioned Video Model.} Existing 2D action-conditioned video models can predict photorealistic futures~\citep{yang2023learning, nvidia2025cosmos, wu2024ivideogpt, jiang2025enerverse, agarwal2026cosmos, liao2026genie}. They have shown promising results in robotic manipulation applications such as data generation~\citep{wang2026interactive, gao2026dreamdojo}, policy evaluation~\citep{guo2026ctrl, quevedo2026worldgym, li2026dworldeval, huang2025vid2world, team2025evaluating}, planning~\cite{zhou2024dino, liang2024dreamitate, chen2025large, hafner2025mastering, hafner2025training, du2023learning, assran2025v}, action prediction~\citep{cheang2024gr, jang2025dreamgen, hu2025video, guo2024prediction, wu2024unleashing, bharadhwaj2025genact}, policy steering~\cite{wu2025foresight, jain2026weaver}, and policy improvement~\citep{yin2026playworld, guo2026vlaw}. However, they do not explicitly represent the 3D world, making it difficult to enforce spatial consistency and model complicated object interactions. Among these works, Interactive World Simulator is the most closely related to ours~\cite{wang2026interactive}. The authors introduce a task-specific 2D action-conditioned video model that supports interactive, stable, and physically plausible long-horizon simulation. However, it predicts in purely pixel space, making prediction from arbitrary viewpoints and spatial consistency challenging. In contrast, our 3D World Simulator explicitly represents the underlying 3D scene, supporting spatially consistent predictions from novel viewpoint. We show that our model can also accurately model challenging object interactions, including cloth folding and part assembly.

\paragraph{3D Dynamics Models.} Many prior works explore modeling object dynamics in the 3D world using various 3D representations, including point cloud~\citep{huang2026pointworld, huang2026soma, bauer2024doughnet}, sparse particles~\citep{zhang2025particle, whitney2024learning, huang2025particleformer, zhang2024adaptigraph, jiang2025phystwin, li2018learning}, neural radiance fields~\citep{driess2023learning}, 3D Gaussians ~\citep{zhang2024dynamic, xie2024physgaussian, shao2025gaussim, abou2024physically, zhang2024physdreamer, lu2024manigaussian, lu2025gwm}, and 3D mesh~\citep{pfaff2020learning}. Some of them focus on learning dynamics of specific types of objects, such as elastoplastic objects~\citep{bauer2024doughnet, shi2023robocook, shi2024robocraft}, cloth~\citep{lin2022learning}, fluid~\citep{sanchez2020learning}. Among these works, PointWorld and PGND are the closest to ours~\citep{huang2026pointworld, zhang2025particle}. Both train 3D world models in point or particle space and show spatially consistent and physically plausible results. However, both use point tracking to provide training supervision. In contrast, our model does not require point-tracking supervision and is transferable across diverse scenes. In addition, they do not focus on photorealistic 2D image rendering, which is essential for visuomotor policy training and evaluation. Instead, our model supports photorealistic 2D rendering from novel viewpoint, including fixed external cameras and moving wrist cameras, enabling scalable data generation and visuomotor policy evaluation.

\paragraph{Multimodal World Models.} Researchers have explored how to incorporate tactile sensing into world models by predicting 2D tactile images alongside 2D visual images~\citep{higuera2026visuotactileworldmodels, zhang2026contactworld, huang2026vitacworld, ma2026feelworld, wu2026inference, tian2019manipulation, lou2026dream, ye2025learning, zang2026tacforesight, yuan2026vtam, wu2026tactile}. In contrast, our model predicts in 3D space with spatial consistency. A closely related work, RoboPack, also models tactile sensing and object interaction dynamics in 3D space to accomplish challenging dense packing tasks~\citep{ai2024robopack}. However, its dynamics model mainly focuses on dense packing tasks and predicts over relatively short horizons.

\section{Method}
\label{sec: method}

% We describe our framework for training an action-conditioned, multimodal, 3D video prediction model. In Section~\ref{sec:vae}, we detail our \textbf{Hybrid Particle-Voxel Autoencoder}. The autoencoder operates on input point clouds $P_{t}$ at time $t$. The encoder $E_{\theta}$ then compresses this point cloud into a compact latent $z_{t}$, from which a decoder $D_{\psi}$ then outputs 3D visual and tactile particles. Each visual particle is attached to a 3D Gaussian that can be rendered from novel viewpoints. Then in Section~\ref{sec:dyn}, we explain our \textbf{3D Latent Dynamics Model}. Given an input latent $z_{t}$, a history with $H$ previous latents $z_{t-H:t-1}$, as well as a corresponding robot action $a_{t}$, the dynamics model $F_{\phi}$ approximates the next latent state $z_{t+1}$ by predicting $\hat{z}_{t+1} = F_{\phi}(z_{t}, z_{t-H:t-1}, a_{t})$.

In this section, we first describe our problem formulation in Section~\ref{sec:prob_form}. Then we will describe how we train our \textbf{hybrid particle-voxel autoencoder} and \textbf{3D latent dynamics model} respectively in Section~\ref{sec:vae} and Section~\ref{sec:dyn}. Lastly, we describe how we rollout autoregressively during the inference time in Section~\ref{sec:inf}

\subsection{Problem Formulation}
\label{sec:prob_form}

We denote the input point clouds at time $t$ as $P_{t}\in \mathbb{R}^{7 \times n}$,
containing $n$ visual and, when available, tactile points. Each point contains its 3D world coordinates, RGB appearance, and an additional channel reserved for tactile information. Our experiments use the FlexiTac~\cite{huang2026flexitac, huang20243d} tactile sensor, but our method is compatible with various types of sensors. Our goal is to build a model that takes in history point clouds $P_{t-H:t}$ and action $a_t$ to predict next $\hat{P}_{t+1}$ minimizing the difference with ground truth ${P}_{t+1}$.

Our model consists of two parts: a hybrid particle-voxel autoencoder and a 3D latent dynamics model. For the autoencoder, the encoder $E_{\theta}$ compresses the point cloud $P_{t}$ into a compact latent $z_{t}$, from which a decoder $D_{\psi}$ then reconstructs the point cloud. To model the dynamics, our 3D latent dynamics model $F_{\phi}$ takes in latent $z_{t}$, last $H$ history latents $z_{t-H:t-1}$, as well as the robot action $a_{t}$ and then predicts $\hat{z}_{t+1} = F_{\phi}(z_{t-H:t}, a_{t})$.

\begin{figure}[t]
\includegraphics[width=\textwidth]{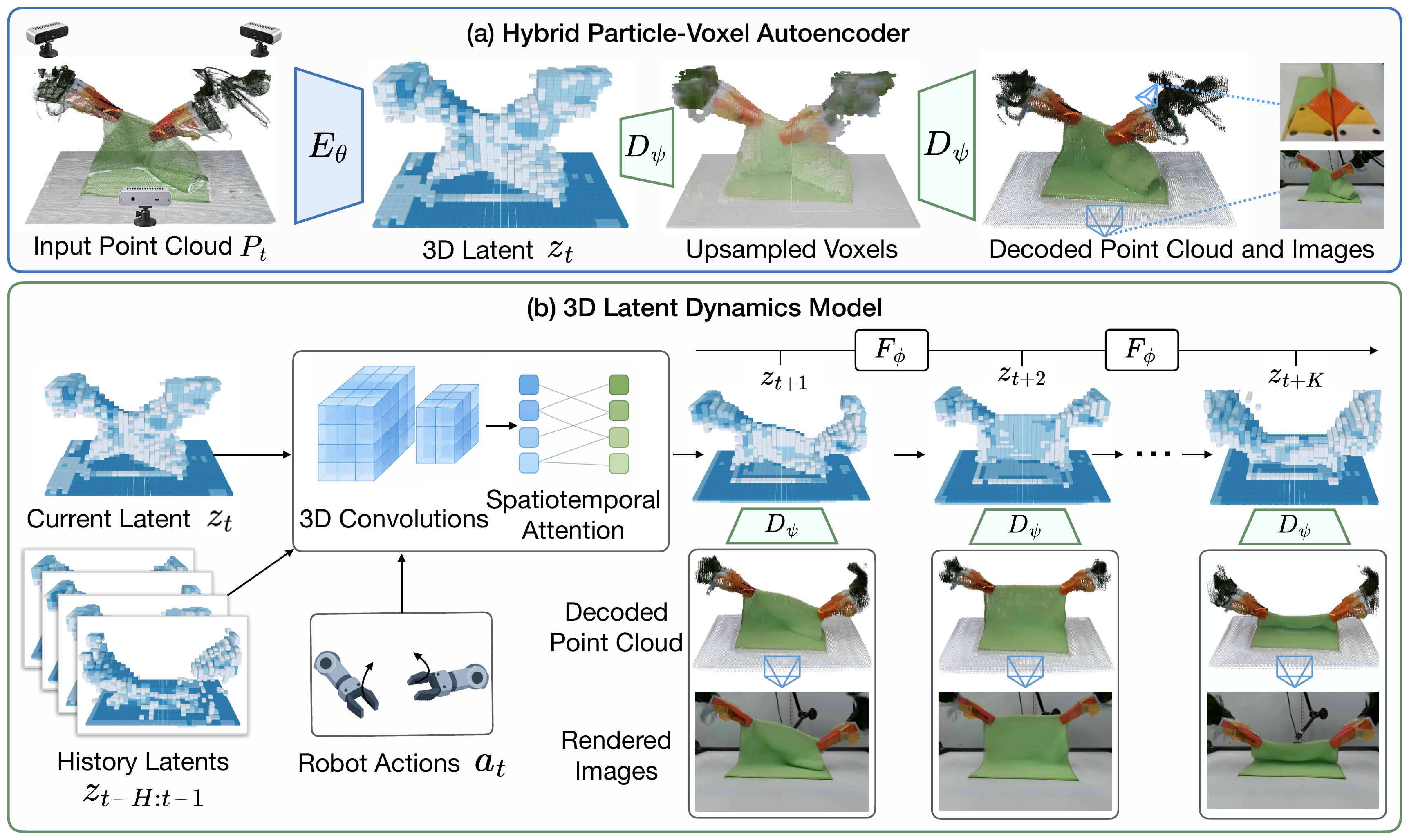}
\vspace{-10pt}
\caption{\small \textbf{Overview of Model Architecture.} (a) We introduce a hybrid particle-voxel autoencoder that compresses an input point cloud into a compact 3D voxel latent. This voxel latent is progressively upsampled into finer voxels, which are then upsampled into dense visual particles and optional tactile particles. Each visual particle is attached to Gaussian parameters that can be rendered at novel viewpoints. (b) Our 3D latent dynamics model then directly operates on voxel latents produced by the autoencoder. Given robot actions, history latents, and a current latent observation, the dynamics model uses 3D convolutions and spatiotemporal attention to autoregressively predict future voxel latents during open-loop rollouts. These latents can then be decoded into point clouds, multi-view renders, and tactile observations. }
\label{fig:method}
\vspace{-5pt}
\end{figure}

\subsection{Hybrid voxel--particle autoencoder}
\label{sec:vae}

% \paragraph{Input Point Cloud.} The autoencoder receives a point cloud
% $P_t = P_t^{\mathrm{visual}} \cup P_t^{\mathrm{tactile}} \in \mathbb{R}^{7 \times n}$ at time $t$,
% containing $n$ visual and, when available, tactile points. Each visual point contains its 3D world coordinates, RGB appearance, and an additional empty feature channel reserved for tactile information. Tactile measurements are represented as 3D points whose positions are obtained from robot forward kinematics, with the continuous tactile sensor value stored in the corresponding feature channel. This provides a common spatial representation in which visual geometry and tactile observations are aligned in the same world coordinate frame. An overview can be found in Figure~\ref{fig:method}.

\paragraph{Sparse voxel encoder.}
We first voxelize the input point cloud $P_{t}$ onto a sparse 3D voxel grid. We aggregate points by averaging the features of points that fall within each occupied voxel, producing features only at occupied voxel locations. The encoder $E_{\theta}$ then applies strided sparse 3D convolutions~\citep{graham2018submanifold} to progressively downsample the voxel input into a compact latent $z_{t}$, avoiding unnecessary computation over empty space. We unit-normalize each latent voxel along the channel dimension before passing the representation to the decoder or latent dynamics model. 

\vspace{-5pt}

\paragraph{Coarse-to-fine particle-voxel decoder.}
As shown in part (a) of Figure~\ref{fig:method}, the decoder $D_{\psi}$ reconstructs the scene from $z_t$ in two stages, upsampling coarse voxels to dense particles. 

\emph{(i) Voxel Upsampling:} the decoder progressively upsamples $z_t$ using transposed 3D convolutions until reaching the desired voxel resolution. At each resolution, the decoder predicts scene occupancy, providing progressively finer estimates of the scene structure at each layer. 

% \emph{(ii) Particle Upsampling:} because voxels provide a coarse representation of scene geometry, we further refine occupied voxels into dense 3D particles. For each predicted occupied voxel, a lightweight MLP predicts continuous particle positions. Subsequent stages then further subdivide and refine these particles, yielding a dense particle representation. Each resulting particle predicts its 3D position and the parameters of a 3D Gaussian, including color, scale, rotation, and opacity, enabling rendering from arbitrary viewpoints. 

\emph{(ii) Particle Upsampling:} to obtain dense 3D particle prediction, we use a lightweight MLP to predict continuous particle positions given each occupied voxel. Each resulting particle predicts its 3D position and the parameters of a 3D Gaussian, including color, scale, rotation, and opacity, enabling rendering from arbitrary viewpoints. 

\paragraph{Training Objective.}

% We train the voxel-particle autoencoder end-to-end with supervision at three levels with regularization: 
We train the voxel-particle autoencoder end-to-end using the following loss:
$\mathcal{L}_{\mathrm{AE}} = \mathcal{L}_{\mathrm{voxel}}
+
\mathcal{L}_{\mathrm{particle}}
+
\mathcal{L}_{\mathrm{render}}
+
\mathcal{L}_{\mathrm{cons}}$.

% \emph{(i) Voxel loss.} We supervise voxel occupancy at each decoder resolution level using a binary cross entropy loss $\mathcal{L}_{\mathrm{voxel}}= \sum_{\ell}\lambda_{\mathrm{occ},\ell}
% \mathcal{L}_{\mathrm{BCE}}^{(\ell)}$, where $\ell$ denotes each level. 

\emph{(i) Voxel loss} $\mathcal{L}_{\mathrm{voxel}}$. To reconstruct the 3D occupancy, we compare predicted voxel occupancy at each decoder resolution level with the ground truth occupancy and sum the loss across all levels.

% \emph{(ii) Particle loss.} We supervise particle geometry using a local bidirectional Chamfer distance (CD)~\citep{fan2017point} between predicted particles and nearby ground-truth points, defined in Equation~\ref{eq: chamfer}. Particle color and tactile values are additionally supervised using $\ell_1$ error on the matched points. 

\emph{(ii) Particle loss} $\mathcal{L}_{\mathrm{particle}}$. To reconstruct accurate particle geometry, we measure the Chamfer distance (CD)~\citep{fan2017point} between predicted particles and ground truth points. Particle color and tactile values are additionally supervised using $\ell_1$ error on the matched points. 

% \emph{(iii) Render loss.} we render the predicted 3D Gaussians from the training camera viewpoints and supervise the resulting RGB-D observations using a combination of $\ell_1$, SSIM~\citep{wang2004image}, and LPIPS~\citep{zhang2018unreasonable} losses on RGB and an $\ell_1$ loss on valid depth pixels, such that $\mathcal{L}_{\mathrm{render}}
% =
% \lambda_{\mathrm{RGB}}
% \left(
% \mathcal{L}_{1}^{\mathrm{RGB}}
% +
% \lambda_{\mathrm{SSIM}}\mathcal{L}_{\mathrm{SSIM}}
% +
% \mathcal{L}_{\mathrm{LPIPS}}
% \right) + \mathcal{L}_{1}^{\mathrm{Depth}}$. 

\emph{(iii) Render loss} $\mathcal{L}_{\mathrm{render}}$. Given the predicted 3D Gaussians, we render RGB-D images from the training camera viewpoints and supervise them using a combination of $\ell_1$, SSIM~\citep{wang2004image}, and LPIPS~\citep{zhang2018unreasonable} losses on RGB and an $\ell_1$ loss on valid depth pixels.

% \emph{(iv) Consistency regularization.} One problem with training latent dynamics models over 3D point cloud data is that point clouds have sampling, size, and ordering variability, leading to inconsistent latents for the same pointcloud. To ensure consistent and stable latents, we use a consistency loss that takes two independent subsamples $P^{(1)}_t, P^{(2)}_t$ of the same pointcloud, and minimizes the $\ell_{2}$ distance between two encoded latents $\mathcal{L}_{\mathrm{cons}} = ||E(P_{t}^{(1)}) - E(P_{t}^{(2)})||^2_{2}$. 

\emph{(iv) Consistency regularization} $\mathcal{L}_{\mathrm{cons}}$. Different subsamples of the same observation can produce inconsistent
latent encodings. We therefore encode two independently subsampled
point clouds and penalize the $\ell_2$ difference between their latent representations.

% , i.e. $\mathcal{L}_{\mathrm{cons}} = ||E_\theta(P_{t}^{(1)}) - E_\theta(P_{t}^{(2)})||^2_{2}$

\subsection{3D Latent Dynamics Model}
\label{sec:dyn}

After training the particle-voxel autoencoder, we learn an action-conditioned dynamics model $F_{\phi}$ over its 3D latent representation. As shown in part (b) of Figure~\ref{fig:method}, at time $t$, given an input latent $z_{t}$, a history with $H$ previous latents $z_{t-H:t-1}$, as well as a corresponding robot action $a_{t}$, the model approximates the next latent state $z_{t+1}$ by predicting $\hat{z}_{t+1} = F_{\phi}(z_{t-H:t}, a_{t})$. The predicted latent can then be decoded into 3D geometry, multi-view visual observations, and tactile signals.

% Our dynamics model consists of a 3D U-Net~\citep{cicek20163d} that processes each input frame with shared 3D convolutional layers. Since attention across 3D voxels is expensive, the U-Net downsamples the latents to a coarser resolution. Once downsampled, we apply interleaved spatial and causal temporal self-attention blocks~\citep{vaswani2017attention} to capture spatial
% and temporal dependencies. A 3D U-Net then upsamples these features back to the input latent resolution. Our approach directly predicts the next latent without any diffusion or iterative sampling. 

% Our dynamics model consists of a 3D U-Net~\citep{cicek20163d} that processes
% each input frame with shared 3D convolutional layers. Since attention over all
% fine-resolution spatiotemporal voxels is expensive, the U-Net first downsamples
% the latent grids to a coarser resolution. We then apply interleaved spatial
% self-attention within each frame and causal temporal self-attention across
% frames~\citep{vaswani2017attention,bertasius2021space} to capture spatial and
% temporal dependencies. The decoder upsamples the resulting features back to
% the input latent resolution. Our model directly predicts the next latent
% without diffusion or iterative sampling.

We instantiate our dynamics model as a 3D U-Net~\citep{cicek20163d}.
We first downsample input latent voxels with 3D convolutional layers and then apply interleaved spatial attention and causal temporal attention~\citep{vaswani2017attention,bertasius2021space}.
Lastly, the decoder upsamples the predicted latent voxels back to the input latent resolution.

\paragraph{Training Objective.}

To improve long-horizon inference stability, we predict $K-$step latents $\hat{z}_{t+1:t+k}$ autoregressively during training and supervise latent predictions using a discounted loss,
% We supervise future latent predictions using a discounted mean-squared error,
\begin{equation}
    \mathcal{L}_{\mathrm{dyn}}
    =
    \frac{1}{K}
    \sum_{k=1}^{K}
    \gamma^{k-1}
    % \operatorname{MSE}
    % \left(
    %     \hat{z}_{t+k},
    %     z_{t+k}
    % \right),
    ||
        \hat{z}_{t+k}-
        z_{t+k}
    ||_2^2,
\end{equation}
% where $K$ denotes the number of supervised autoregressive predictions and
% $\gamma$ is the discount factor.
where $\gamma$ is the discount factor.
For stable training, we use a curriculum that gradually increases $K$ during training, similar in spirit to scheduled sampling~\citep{bengio2015scheduled}. To further improve the long-horizon inference stability, we add randomly sampled Gaussian noise to each history latent during training. More details of model and training specifications are in Appendix~\ref{appendix:model-details}.
% To improve long-horizon fidelity, we add randomly sampled Gaussian noise to
% each history latent during training. We further expose the model to its own
% prediction errors by feeding autoregressively generated latents back into the
% history. We use a curriculum that gradually increases the number of
% autoregressive feedback steps during training, similar in spirit to scheduled
% sampling~\citep{bengio2015scheduled}. We provide detailed model and training
% specifications in Appendix~\ref{appendix:model-details}.

% We supervise future latent predictions using mean-squared error. To improve long horizon fidelity, we add randomly sampled Gaussian noise to each history latent during training. We also train the model to be robust to its own predictions and use autoregressively generated predictions as input history latents during training. We utilize a curriculum that gradually increases the number of autoregressive predictions $k$ fed back into the model during training, up to $K$ rollout steps. Thus, the dynamics loss is $\mathcal{L}_{\mathrm{dyn}} = \sum_{k=1}^{K}\left \|\hat{z}_{t+k} - z_{t+k} \right\|_{2}^{2}$. We include detailed model training and loss specifics for both the autoencoder and dynamics model in Appendix~\ref{appendix:model-details}. 

The hybrid particle-voxel autoencoder and 3D latent dynamics model takes around 16 hours each on an RTX 5090 GPU. 

\subsection{Unified composition of simulation "backbones"}
\label{sec:composition}
While our model is limited to a specific workspace, to enable larger scale scene simulation, our model can be composed with various 3D Gaussian assets. By composing the raw autoencoder output with diverse static background made up of 3D Gaussians, we can further simulate various types of scenes. We can further utilize kinematic 3D Gaussian robot models to allow for more complete scene simulation, as robot arms may fall outside the specificied workspace. Although our model does not natively depend on object-specific representations, it can be finetuned to output labels to separate specific scene components like the robot arm, object, and table. Given this separability, we can further insert different robot arms and kinematic models not seen during training, as well as diverse backgrounds.

\subsection{Inference}
\label{sec:inf}
At inference time, we roll out the world model autoregressively in open loop. Starting from an initial encoded 3D latent $z_1$, the dynamics model predicts the next 3D latent conditioned on the robot action, and each predicted latent is fed back as input for subsequent steps. Each predicted latent can be decoded into visual and tactile point clouds, as well as 3D Gaussians that can be rendered from arbitrary viewpoints. To render wrist camera views during open-loop rollouts, we fine-tune a prediction head on the decoder.
% For viewpoints that were not covered by cameras during training as well as scene points that fall outside the voxel grid, our method can incorporate static backgrounds or kinematic robot Gaussian Splats to produce renders at a larger scene scale.

\section{Experiments}
\label{sec:experiments}

\begin{figure*}[t]
\centering
\includegraphics[width=\textwidth]{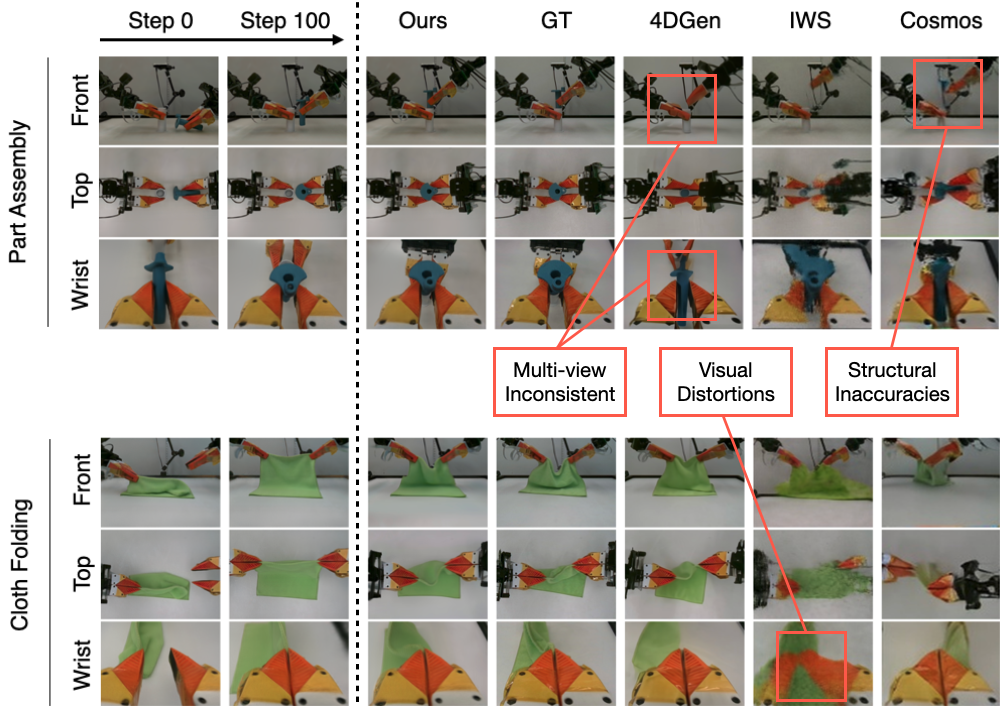}
\caption{\small \textbf{Qualitative comparison of open-loop multi-view predictions.} (Left) Visualization of 3D World Simulator multi-view renders from open-loop rollouts on the Part Assembly and Cloth Folding task. Our model maintains multi-view consistent predictions with strong visual fidelity. (Right) Comparison between our method and baselines on frame 200 of an open-loop rollout. Other baselines display inconsistent predictions between viewpoints, visual distortions, and object and robot structural inaccuracies. }
\label{fig: multiview-qualitative}
\end{figure*}

\begin{figure*}[t]
\centering
\includegraphics[width=\textwidth]{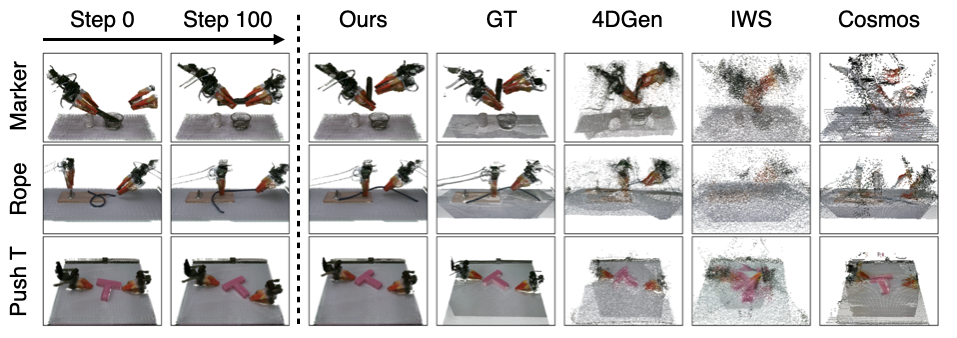}
\caption{\small \textbf{Qualitative comparison of open-loop point cloud predictions.} (Left) Open-loop rollout point cloud visualizations for our method on the Marker Insertion, Rope Threading, and T Pushing task. (Right) Our method maintains consistent point clouds close to ground truth, while other video based methods show misaligned and inaccurate point clouds.}
\label{fig: pointcloud-qualitative}
\end{figure*}

Through our experiments, we aim to address three questions: (1) How does the 3D World Simulator compare to baselines in terms of long-horizon prediction of 3D geometry, multi-view consistent renders and tactile signals? (2) Can the 3D World Simulator be used to train and evaluate imitation learning policies, especially those that utilize wrist-mounted cameras, tactile signals, or multi-view observations? (3) Can the 3D World Simulator scale from individual tasks to diverse objects and interactions when trained on large-scale data, and be used for downstream planning with unseen objects and robot arms? 

\begin{figure*}[t]
\centering
\includegraphics[width=\textwidth]{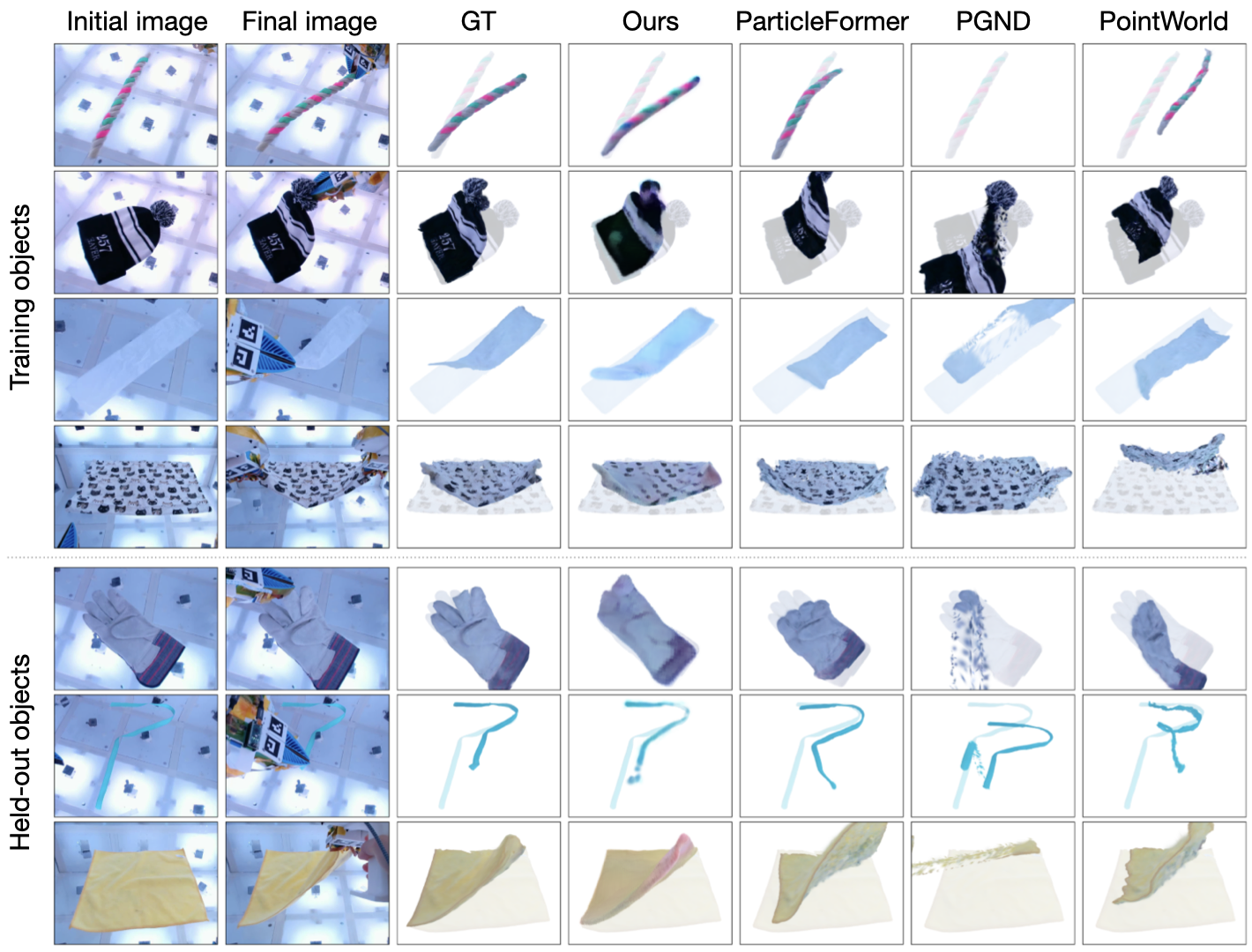}
\caption{\small \textbf{Qualitative Comparison on Deform360.} The left two columns show the starting and end position for each object, while the right columns show rendered Gaussians produced by each method. On both objects seen during training and objects not seen during training, our method has better visually matching geometry compared to baseline methods. }
\label{fig: qualitative-deform360}
\end{figure*}
\vspace{-5pt}

\subsection{Experiment Setup}
\label{subsection: experiment-and-world-model-setup}
\noindent \textbf{ALOHA Task Setup.} We conduct data collection and experimentation using an ALOHA Bimanual Robot~\citep{zhao2023act}, paired with FlexiTac~\citep{huang2026flexitac} tactile sensors on each fingertip. We include six tasks that cover manipulation of deformable, rigid, and articulated objects: \textit{Part Assembly, Cloth Folding, T Pushing, Box Packing, Rope Threading, and Marker Insertion}. Detailed descriptions and visualizations of each task setup can be found in Appendix~\ref{appendix:aloha-task-setup}.  We collect data via a combination of kinematic devices and keyboard inputs to teleoperate the robot, resulting in around 200 episodes and 2 hours of data for each task.

\noindent \textbf{Deform360 Dataset.} To investigate the scalability of 3D World Simulator, we utilize Deform 360~\citep{li2026deform360}, a visuo-tactile dataset featuring 198 daily-life objects, 1,980 interaction sequences, and over 215 hours of observations from 41 surround-view cameras and bimanual tactile grippers. Since Deform360 only provides object-centric depth information, we train our world model purely on the object particles unlike the ALOHA setup where we reconstruct the robot and environment surroundings. 

\subsection{2D Video Prediction and 3D Particle Baseline Comparisons}

\begin{table*}[!t]
\centering
\small
\resizebox{0.86\textwidth}{!}{%
\begin{tabular}{@{}c@{\hspace{12pt}}lccccccc@{}}
\toprule
& & \multicolumn{2}{c}{Image} & \multicolumn{2}{c}{Depth} & \multicolumn{2}{c}{Geometry} & \multicolumn{1}{c}{Tactile} \\
\cmidrule(lr){3-4} \cmidrule(lr){5-6} \cmidrule(lr){7-8} \cmidrule(lr){9-9}
& Method & PSNR $\uparrow$ & LPIPS $\downarrow$ & Abs Rel $\downarrow$ & $\delta < 1.25$ $\uparrow$ & CD $\downarrow$ & Obj. CD $\downarrow$ & L1 $\downarrow$ \\
\midrule
\multirow{3}{*}{\rotatebox{90}{\scriptsize \textbf{Video}}} & Cosmos & 18.51 & 0.441 & 0.147 & 0.833 & 12.43 & 16.64 & 0.135 \\
& Robot4DGen & 20.59 & 0.340 & 0.172 & 0.790 & 10.43 & 15.97 & 0.128 \\
& IWS & 12.59 & 0.562 & 0.163 & 0.791 & 10.81 & 17.69 & 0.207 \\
\midrule
\multirow{3}{*}{\rotatebox{90}{\scriptsize \textbf{Particle}}} & PGND & -- & -- & -- & -- & 17.11 & 25.48 & 0.208 \\
& PointWorld & -- & -- & -- & -- & 15.32 & 21.74 & 0.148 \\
& ParticleFormer & -- & -- & -- & -- & 13.78 & 20.23 & 0.151 \\
\midrule
& \textbf{Ours} & \textbf{23.24} & \textbf{0.240} & \textbf{0.097} & \textbf{0.913} & \textbf{7.72} & \textbf{11.52} & \textbf{0.081} \\
\bottomrule
\end{tabular}%
}
\caption{\small \textbf{Quantitative Comparison on ALOHA tasks} The 3D World Simulator outperforms both 2D video and 3D particle baselines in terms of visual, depth, geometry, and tactile metrics. Results are averaged over 6 tasks and averaged over a horizon of 200 steps. While using explicit 3D, particle baselines degrade over long horizons, particularly around occlusions.  }
\label{table: quantitative-forward-prediction}
\end{table*}
% \begin{table*}[!t]
% \centering
% \small
% \resizebox{0.8\textwidth}{!}{%
% \begin{tabular}{@{}c@{\hspace{12pt}}lcccccc@{}}
% \toprule
% & & \multicolumn{3}{c}{Training objects} & \multicolumn{3}{c}{Held-out objects} \\
% \cmidrule(lr){3-5} \cmidrule(lr){6-8}
% & Method & PSNR $\uparrow$ & LPIPS $\downarrow$ & CD $\downarrow$ & PSNR $\uparrow$ & LPIPS $\downarrow$ & CD $\downarrow$ \\
% \midrule
% \multirow{3}{*}{\rotatebox{90}{\scriptsize \textbf{Particle}}} & PGND & 14.9 & 0.41 & 18.6 & 12.3 & 0.50 & 25.2 \\
% & PointWorld & 16.8 & 0.34 & 12.4 & 15.0 & 0.58 & 14.5 \\
% & ParticleFormer & 18.0 & 0.30 & 10.5 & \textbf{15.6} & \textbf{0.31} & 13.4 \\
% \midrule
% & \textbf{Ours} & \textbf{18.6} & \textbf{0.26} & \textbf{9.6} & 14.8 & 0.43 & \textbf{12.1} \\
% \bottomrule
% \end{tabular}%
% }
% % \vspace{-5pt}
% \caption{\small \textbf{Quantitative Comparison on Deform360.} Results are averaged over 150 step open-loop rollouts. On objects seen during training, 3D World Simulator outperforms all methods on both image and geometry metrics. On held-out objects, our method retains better geometry but worse rendering quality. This advantage can be explained by particle methods outputting tracks on fitted Gaussian Splats, while our method generates appearance from on unseen objects. }
% \label{tab:quantitative-deform360}
% \vspace{-10pt}
% \end{table*}
% I think we want to move this table to appendix as the result is somewhat mixed?

\noindent \textbf{Baselines and Metrics.}
Across the six ALOHA tasks in Section~\ref{subsection: experiment-and-world-model-setup}, we evaluate 3D World Simulator against both 2D Video Prediction and 3D Particle Baselines on metrics across pixel reconstruction (PSNR and LPIPS), depth prediction (Abs Rel, $\delta < 1.25$~\citep{eigen2014depth}), point cloud reconstruction (Chamfer distance on scene and object points), and tactile signal prediction metrics (L1).  We evaluate long horizon prediction by averaging metrics across an open-loop rollout of 200 frames. On the Deform360 dataset, we only compare to 3D particle baselines since per-view scene depth is not contained in the dataset. 

For representative 2D video prediction baselines, we include Robot4DGen~\citep{liu2026geometry}, Cosmos~\citep{nvidia2025cosmos}, and Interactive World Simulator (IWS)~\citep{wang2026interactive}. For representative 3D particle baselines, we include ParticleFormer~\citep{huang2025particleformer}, PointWorld~\citep{huang2026pointworld}, and Particle-Grid Neural Dynamics (PGND)~\citep{zhang2025particle}.

\noindent \textbf{ALOHA Task Results.} Quantitative results in Table~\ref{table: quantitative-forward-prediction} show that 3D World Simulator outperforms all 2D and 3D baselines across all metrics, including pixel reconstruction, depth prediction, point cloud reconstruction, and tactile signal prediction. Qualitative results in Figure~\ref{fig: multiview-qualitative} show that 3D World Simulator produces multi-view consistent renders over long horizons, while other 2D video prediction baselines show inconsistency between views, inaccurate dynamics, and accumulated artifacts.  Qualitative point cloud reconstructions in Figure~\ref{fig: pointcloud-qualitative} show that baseline predictions yield inconsistent point clouds.

\begin{figure*}[t]
\centering
\includegraphics[width=0.8\textwidth]{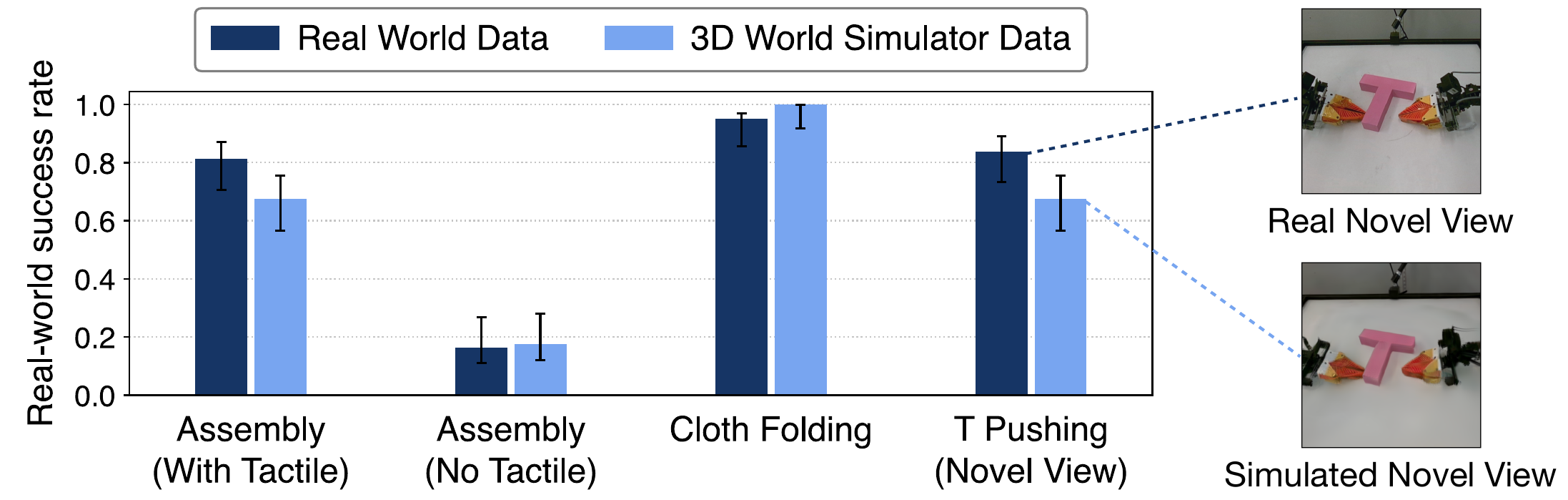}
\vspace{-5pt}
\caption{\small \textbf{Policy Training on 3D World Simulator and Real World Data.} Imitation learning policies trained using purely 3D World Simulator generated data show comparable performance to policies trained using real world data, across visuo-tactile, multi-view, and novel view observations generated by our model.}
\label{fig: policy-training}
\vspace{-12pt}
\end{figure*}

\begin{figure*}[t]
\centering
\vspace{8pt}
\includegraphics[width=\textwidth]{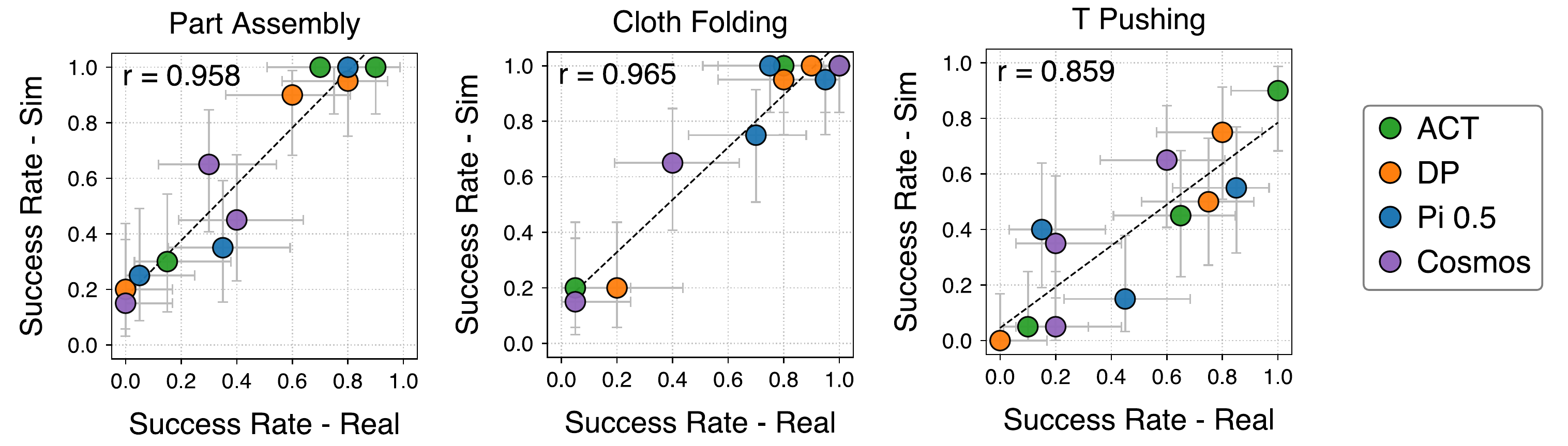}
\vspace{-18pt}
\caption{\small \textbf{Correlation between simulation and real-world policy performance.} Policies evaluated in 3D World Simulator and the real world show strong correlation across all three tasks. }
\label{fig: correlation-plots}
\vspace{-10pt}
\end{figure*}

\noindent \textbf{Deform360 Results.} 
 Table~\ref{tab:quantitative-deform360} shows that 3D World Simulator outperforms 3D particle baselines in terms of image and geometry metrics on objects seen during training. On objects held-out during training, 3D World Simulator produces better geometric predictions but has worse results on image metrics. This discrepancy can be explained by the fact that the particle baselines output point tracks on prefitted Gaussian Splats, while ours generates appearance from scratch. Qualitative results on the Deform360 dataset in Figure~\ref{fig: qualitative-deform360} show that our method is able to generate realistic dynamics and appearance on both seen object and held-out during training. 

\subsection{Data Generation for Policy Training}
\label{subsection: policy-training}

We further evaluate whether 3D World Simulator can act as a scalable data generator, particularly for policies that rely on wrist-mounted cameras, tactile signals, or novel views not captured by the cameras during data collection. We evaluate four imitation learning policies:
ACT~\citep{zhao2023act},
Diffusion Policy (DP)~\citep{chi2023diffusion},
$\pi_{0.5}$~\citep{black2025pi05}, and
Cosmos Policy~\citep{kim2026cosmos}. We consider three tasks: \textit{Part Assembly, Cloth Folding, and T-Pushing}. For each task, we collect 100 demonstrations in the real world and in 3D World Simulator as well. We report the mean success rate across the four policy architectures. Detailed descriptions, visualizations, and evaluation criteria are described in Appendix~\ref{appendix:policy-training}. Each task isolates a different capability of the simulator. \textit{\textbf{Part Assembly}} evaluates visuo-tactile policy training, \textit{\textbf{T-Pushing}} evaluates policy learning from novel viewpoints, and \textit{\textbf{Cloth Folding}} evaluates multi-view consistency using front and wrist-mounted camera observations.  As shown in Figure~\ref{fig: policy-training}, \textit{Part Assembly} policies trained on simulated visuo-tactile data achieve performance comparable to those trained on real data and substantially outperform vision-only policies trained on real data, showing the utility of tactile signals generated by our model.  Furthermore, \textit{Cloth Folding } and \textit{T-Pushing} results indicate the utility of our simulator for multi-view and novel view observations.

\vspace{-4pt}
\subsection{Faithful Policy Evaluation}
\label{subsection:policy-evaluation}
\vspace{-4pt}

To evaluate whether 3D World Simulator can serve as a faithful proxy for real-world policy evaluation, we examine whether imitation policies show positive correlation when evaluated within our simulator and the real world. We train four imitation learning policies: ACT, DP, $\pi_{0.5}$, and Cosmos Policy, and we utilize the same three tasks from policy training in Section~\ref{subsection: policy-training}, \textit{Part Assembly, Cloth Folding, and T-Pushing}. We sample $20$ initial configurations for each task.
\begin{figure*}[t]
\centering
\vspace{5pt}
\includegraphics[width=\textwidth]{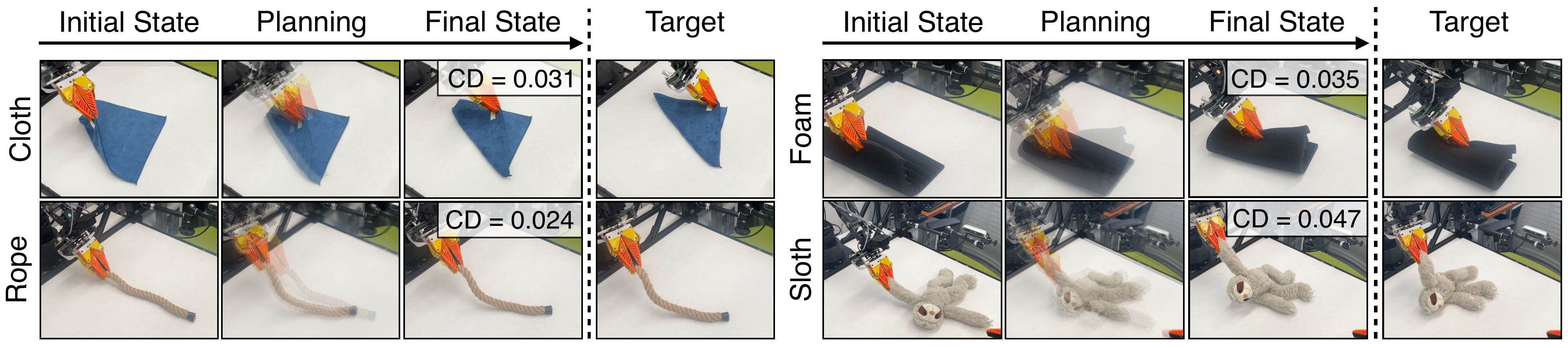}
\vspace{-15pt}
\caption{\small \textbf{Model-based planning on unseen objects in the real world.} Utilizing the pre-trained 3D World Simulator on Deform360, our model generalizes to downstream model-based planning on four unseen objects and an unseen ALOHA robot arm. }
\label{fig: planning}

\end{figure*} Figure~\ref{fig: correlation-plots} shows that across all three tasks and four policies, real-world policy performance is highly correlated with performance in the 3D World Simulator. The $x$-axis denotes the real-world success rate while the $y$-axis denotes the success rate within the world simulator. Each point corresponds to a policy checkpoint at a different training step, while colors denote the policy architecture. Error bars are indicated through Clopper-Pearson confidence intervals, obtained through $20$ evaluations per policy both in simulation and the real world.

\subsection{Model-Based Planning on Unseen Objects}
We utilize our pre-trained model on the large-scale Deform360 dataset for downstream model-based planning. We utilize an unseen ALOHA single-arm setup with four unseen objects: cloth, rope, foam, and sloth. We utilize Model Predictive Control, using an off-the-shelf implementation of MPPI~\citep{williams2017model} from PGND~\citep{zhang2025particle}. Figure~\ref{fig: planning} shows qualitatively and quantitatively that our pre-trained 3D World Simulator can achieve zero-shot planning results.

\section{Conclusion}
In this work, we introduced 3D World Simulator, an action-conditioned multimodal world modeling framework that predicts in a unified 3D representation space, creating a common interface to generate 3D particles, multi-view consistent rendered RGB, and tactile predictions. Our unified model enables a range of applications. 3D World Simulator enables the generation of policy training data for visuo-tactile policies, as well as policies that rely on multi-view, novel view, and wrist-mounted camera view observations. We can further evaluate these policies using our model, demonstrating strong correlation between simulation and real world rollouts. When scaled to large scale datasets, 3D World Simulator enables downstream model-based planning on unseen objects and robot configurations. In future work, we plan to extend 3D World Simulator to more diverse and challenging environments, extending its general applicability. Additionally, we further aim to extend 3D World Simulator to larger scale scenes involving multi-room environments and mobile manipulation, enabling increased real world applications.

\subsection*{Acknowledgement}
This work was partially supported by Analog Devices, the Toyota Research Institute, and NSF Award \#2409661. This article solely reflects the opinions and conclusions of its authors and should not be interpreted as necessarily representing the official policies, either expressed or implied, of the sponsors.

% \subsection*{Reproducibility Statement}
% We provide our code so our results can be easily reproduced. 

\bibliography{iclr2027_conference}

@inproceedings{liu2026geometry,
  title={Geometry-aware 4D Video Generation for Robot Manipulation},
  author={Liu, Zeyi and Li, Shuang and Cousineau, Eric and Feng, Siyuan and Burchfiel, Benjamin and Song, Shuran},
  booktitle={International Conference on Learning Representations (ICLR)},
  year={2026}
}

@article{nvidia2025cosmos,
  title={Cosmos World Foundation Model Platform for Physical AI},
  author={{NVIDIA} and Agarwal, Niket and Ali, Arslan and Bala, Maciej and others},
  journal={arXiv preprint arXiv:2501.03575},
  year={2025}
}

@article{kerbl2023gaussian,
  title={3D Gaussian Splatting for Real-Time Radiance Field Rendering},
  author={Kerbl, Bernhard and Kopanas, Georgios and Leimk{\"u}hler, Thomas and Drettakis, George},
  journal={ACM Transactions on Graphics},
  volume={42},
  number={4},
  pages={1--14},
  year={2023}
}

@inproceedings{zhang2018unreasonable,
  title={The Unreasonable Effectiveness of Deep Features as a Perceptual Metric},
  author={Zhang, Richard and Isola, Phillip and Efros, Alexei A. and Shechtman, Eli and Wang, Oliver},
  booktitle={Proceedings of the IEEE Conference on Computer Vision and Pattern Recognition (CVPR)},
  pages={586--595},
  year={2018}
}

@article{wang2004image,
  title={Image Quality Assessment: From Error Visibility to Structural Similarity},
  author={Wang, Zhou and Bovik, Alan C. and Sheikh, Hamid R. and Simoncelli, Eero P.},
  journal={IEEE Transactions on Image Processing},
  volume={13},
  number={4},
  pages={600--612},
  year={2004},
  doi={10.1109/TIP.2003.819861}
}

@inproceedings{li2026deform360,
  title={Deform360: A Massive Multi-view Visuotactile Dataset for Deformable World Models},
  author={Li, Hongyu and Fu, Wanjia and Cong, Xiaoyan and Li, Zekun and Huang, Binghao and Jiang, Hanxiao and He, Xintong and Liang, Yiqing and Fu, Rao and Lu, Tao and Sridhar, Srinath and Smith, Kevin A. and Konidaris, George and Li, Yunzhu},
  booktitle={European Conference on Computer Vision (ECCV)},
  year={2026}
}

@inproceedings{zhang2025particle,
  title={Particle-Grid Neural Dynamics for Learning Deformable Object Models from RGB-D Videos},
  author={Zhang, Kaifeng and Li, Baoyu and Hauser, Kris and Li, Yunzhu},
  booktitle={Proceedings of Robotics: Science and Systems (RSS)},
  year={2025}
}

@inproceedings{huang2026pointworld,
  title={PointWorld: Scaling 3D World Models for In-The-Wild Robotic Manipulation},
  author={Huang, Wenlong and Chao, Yu-Wei and Mousavian, Arsalan and Liu, Ming-Yu and Fox, Dieter and Mo, Kaichun and Li, Fei-Fei},
  booktitle={Proceedings of the IEEE/CVF Conference on Computer Vision and Pattern Recognition (CVPR)},
  year={2026}
}

@inproceedings{wang2026interactive,
  title={Interactive World Simulator for Robot Policy Training and Evaluation},
  author={Wang, Yixuan and Syed, Rhythm and Wu, Fangyu and Zhang, Mengchao and Onol, Aykut and Barreiros, Jose and Nayyeri, Hooshang and Dear, Tony and Zhang, Huan and Li, Yunzhu},
  booktitle={Proceedings of Robotics: Science and Systems (RSS)},
  year={2026}
}

@inproceedings{huang2025particleformer,
  title={ParticleFormer: A 3D Point Cloud World Model for Multi-Object, Multi-Material Robotic Manipulation},
  author={Huang, Suning and Chen, Qianzhong and Zhang, Xiaohan and Sun, Jiankai and Schwager, Mac},
  booktitle={Proceedings of the 9th Conference on Robot Learning},
  pages={4941--4957},
  year={2025},
  volume={305},
  series={Proceedings of Machine Learning Research}
}

@inproceedings{wen2025foundationstereo,
  title={FoundationStereo: Zero-Shot Stereo Matching},
  author={Wen, Bowen and Trepte, Matthew and Aribido, Joseph and Kautz, Jan and Gallo, Orazio and Birchfield, Stan},
  booktitle={Proceedings of the IEEE/CVF Conference on Computer Vision and Pattern Recognition (CVPR)},
  pages={5249--5260},
  year={2025}
}

@inproceedings{guo2026ctrl,
  title={Ctrl-world: A controllable generative world model for robot manipulation},
  author={Guo, Yanjiang and Shi, Lucy and Chen, Jianyu and Finn, Chelsea},
  booktitle={International Conference on Learning Representations},
  volume={2026},
  pages={6121--6138},
  year={2026}
}

@misc{huang2026flexitac,
      title={FlexiTac: A Low-Cost, Open-Source, Scalable Tactile Sensing Solution for Robotic Systems}, 
      author={Binghao Huang and Yunzhu Li},
      year={2026},
      eprint={2604.28156},
      archivePrefix={arXiv},
      primaryClass={cs.RO},
      url={https://arxiv.org/abs/2604.28156}, 
}

@inproceedings{quevedo2026worldgym,
  title={Worldgym: World model as an environment for policy evaluation},
  author={Quevedo, Julian and Sharma, Ansh Kumar and Sun, Yixiang and Suryavanshi, Varad and Liang, Percy and Yang, Sherry},
  booktitle={International Conference on Learning Representations},
  volume={2026},
  pages={78932--78957},
  year={2026}
}

@article{yang2023learning,
  title={Learning interactive real-world simulators},
  author={Yang, Sherry and Du, Yilun and Ghasemipour, Kamyar and Tompson, Jonathan and Kaelbling, Leslie and Schuurmans, Dale and Abbeel, Pieter},
  journal={arXiv preprint arXiv:2310.06114},
  year={2023}
}

@article{wu2024ivideogpt,
  title={ivideogpt: Interactive videogpts are scalable world models},
  author={Wu, Jialong and Yin, Shaofeng and Feng, Ningya and He, Xu and Li, Dong and Hao, Jianye and Long, Mingsheng},
  journal={Advances in Neural Information Processing Systems},
  volume={37},
  pages={68082--68119},
  year={2024}
}

@article{zhou2024dino,
  title={Dino-wm: World models on pre-trained visual features enable zero-shot planning},
  author={Zhou, Gaoyue and Pan, Hengkai and LeCun, Yann and Pinto, Lerrel},
  journal={arXiv preprint arXiv:2411.04983},
  year={2024}
}

@inproceedings{lu2025gwm,
  title={Gwm: Towards scalable gaussian world models for robotic manipulation},
  author={Lu, Guanxing and Jia, Baoxiong and Li, Puhao and Chen, Yixin and Wang, Ziwei and Tang, Yansong and Huang, Siyuan},
  booktitle={2025 IEEE/CVF International Conference on Computer Vision (ICCV)},
  pages={9263--9274},
  year={2025},
  organization={IEEE}
}

@inproceedings{lu2024manigaussian,
  title={Manigaussian: Dynamic gaussian splatting for multi-task robotic manipulation},
  author={Lu, Guanxing and Zhang, Shiyi and Wang, Ziwei and Liu, Changliu and Lu, Jiwen and Tang, Yansong},
  booktitle={European Conference on Computer Vision},
  pages={349--366},
  year={2024},
  organization={Springer}
}

@misc{higuera2026visuotactileworldmodels,
      title={Visuo-Tactile World Models}, 
      author={Carolina Higuera and Sergio Arnaud and Byron Boots and Mustafa Mukadam and Francois Robert Hogan and Franziska Meier},
      year={2026},
      eprint={2602.06001},
      archivePrefix={arXiv},
      url={https://arxiv.org/abs/2602.06001}, 
}

@article{gao2026dreamdojo,
  title={Dreamdojo: A generalist robot world model from large-scale human videos},
  author={Gao, Shenyuan and Liang, William and Zheng, Kaiyuan and Malik, Ayaan and Ye, Seonghyeon and Yu, Sihyun and Tseng, Wei-Cheng and Dong, Yuzhu and Mo, Kaichun and Lin, Chen-Hsuan and others},
  journal={arXiv preprint arXiv:2602.06949},
  year={2026}
}

@inproceedings{liao2026genie,
  title={Genie envisioner: A unified world foundation platform for robotic manipulation},
  author={Liao, Yue and Zhou, Pengfei and Huang, Siyuan and Yang, Donglin and Chen, Shengcong and Jiang, Yuxin and Hu, Yue and Liu, Si and Luo, Jianlan and Chen, Liliang and others},
  booktitle={International Conference on Learning Representations},
  volume={2026},
  pages={88446--88463},
  year={2026}
}

@article{huang2025vid2world,
  title={Vid2world: Crafting video diffusion models to interactive world models},
  author={Huang, Siqiao and Wu, Jialong and Zhou, Qixing and Miao, Shangchen and Long, Mingsheng},
  journal={arXiv preprint arXiv:2505.14357},
  year={2025}
}

@inproceedings{
guo2026vlaw,
title={{VLAW}: Iterative Co-Improvement of Vision-Language-Action Policy and World Model},
author={Yanjiang Guo and Tony Lee and Lucy Xiaoyang Shi and Jianyu Chen and Percy Liang and Chelsea Finn},
booktitle={Forty-third International Conference on Machine Learning},
year={2026},
url={https://openreview.net/forum?id=68zOaa2gOf}
}

@inproceedings{zhao2023act,
  title={Learning Fine-Grained Bimanual Manipulation with Low-Cost Hardware},
  author={Zhao, Tony Z. and Kumar, Vikash and Levine, Sergey and Finn, Chelsea},
  booktitle={Proceedings of Robotics: Science and Systems (RSS)},
  year={2023}
}

@inproceedings{chi2023diffusion,
  title={Diffusion Policy: Visuomotor Policy Learning via Action Diffusion},
  author={Chi, Cheng and Feng, Siyuan and Du, Yilun and Xu, Zhenjia and Cousineau, Eric and Burchfiel, Benjamin C. M. and Song, Shuran},
  booktitle={Proceedings of Robotics: Science and Systems (RSS)},
  year={2023}
}

@inproceedings{black2025pi05,
  title={{$\pi_{0.5}$}: A Vision-Language-Action Model with Open-World Generalization},
  author={Black, Kevin and Brown, Noah and Darpinian, James and Dhabalia, Karan and Driess, Danny and Esmail, Adnan and Equi, Michael Robert and Finn, Chelsea and Fusai, Niccolo and Galliker, Manuel Y. and Ghosh, Dibya and Groom, Lachy and Hausman, Karol and Ichter, Brian and Jakubczak, Szymon and Jones, Tim and Ke, Liyiming and LeBlanc, Devin and Levine, Sergey and Li-Bell, Adrian and Mothukuri, Mohith and Nair, Suraj and Pertsch, Karl and Ren, Allen Z. and Shi, Lucy Xiaoyang and Smith, Laura and Springenberg, Jost Tobias and Stachowicz, Kyle and Tanner, James and Vuong, Quan and Walke, Homer and Walling, Anna and Wang, Haohuan and Yu, Lili and Zhilinsky, Ury},
  booktitle={Proceedings of The 9th Conference on Robot Learning (CoRL)},
  volume={305},
  pages={17--40},
  year={2025}
}

@article{wu2025foresight,
  title={From foresight to forethought: Vlm-in-the-loop policy steering via latent alignment},
  author={Wu, Yilin and Tian, Ran and Swamy, Gokul and Bajcsy, Andrea},
  journal={arXiv preprint arXiv:2502.01828},
  year={2025}
}

@inproceedings{
jang2025dreamgen,
title={DreamGen: Unlocking Generalization in Robot Learning through Video World Models},
author={Joel Jang and Seonghyeon Ye and Zongyu Lin and Jiannan Xiang and Johan Bjorck and Yu Fang and Fengyuan Hu and Spencer Huang and Kaushil Kundalia and Yen-Chen Lin and Lo{\"\i}c Magne and Ajay Mandlekar and Avnish Narayan and You Liang Tan and Guanzhi Wang and Jing Wang and Qi Wang and Yinzhen Xu and Xiaohui Zeng and Kaiyuan Zheng and Ruijie Zheng and Ming-Yu Liu and Luke Zettlemoyer and Dieter Fox and Jan Kautz and Scott Reed and Yuke Zhu and Linxi Fan},
booktitle={9th Annual Conference on Robot Learning},
year={2025},
url={https://openreview.net/forum?id=3CnxNqmklv}
}

@inproceedings{
hu2025video,
title={Video Prediction Policy: A Generalist Robot Policy with Predictive Visual Representations},
author={Yucheng Hu and Yanjiang Guo and Pengchao Wang and Xiaoyu Chen and Yen-Jen Wang and Jianke Zhang and Koushil Sreenath and Chaochao Lu and Jianyu Chen},
booktitle={Forty-second International Conference on Machine Learning},
year={2025},
url={https://openreview.net/forum?id=c0dhw1du33}
}

@article{guo2024prediction,
  title={Prediction with action: Visual policy learning via joint denoising process},
  author={Guo, Yanjiang and Hu, Yucheng and Zhang, Jianke and Wang, Yen-Jen and Chen, Xiaoyu and Lu, Chaochao and Chen, Jianyu},
  journal={Advances in Neural Information Processing Systems},
  volume={37},
  pages={112386--112410},
  year={2024}
}

@inproceedings{wu2024unleashing,
  title={Unleashing large-scale video generative pre-training for visual robot manipulation},
  author={Wu, Hongtao and Jing, Ya and Cheang, Chilam and Chen, Guangzeng and Xu, Jiafeng and Li, Xinghang and Liu, Minghuan and Li, Hang and Kong, Tao},
  booktitle={International Conference on Learning Representations},
  volume={2024},
  pages={10641--10662},
  year={2024}
}

@article{liang2024dreamitate,
  title={Dreamitate: Real-world visuomotor policy learning via video generation},
  author={Liang, Junbang and Liu, Ruoshi and Ozguroglu, Ege and Sudhakar, Sruthi and Dave, Achal and Tokmakov, Pavel and Song, Shuran and Vondrick, Carl},
  journal={arXiv preprint arXiv:2406.16862},
  year={2024}
}

@inproceedings{
bharadhwaj2025genact,
title={Gen2Act: Human Video Generation in Novel Scenarios enables Generalizable Robot Manipulation},
author={Homanga Bharadhwaj and Debidatta Dwibedi and Abhinav Gupta and Shubham Tulsiani and Carl Doersch and Ted Xiao and Dhruv Shah and Fei Xia and Dorsa Sadigh and Sean Kirmani},
booktitle={9th Annual Conference on Robot Learning},
year={2025},
url={https://openreview.net/forum?id=HprBJupvvM}
}

@article{du2023learning,
  title={Learning universal policies via text-guided video generation},
  author={Du, Yilun and Yang, Sherry and Dai, Bo and Dai, Hanjun and Nachum, Ofir and Tenenbaum, Josh and Schuurmans, Dale and Abbeel, Pieter},
  journal={Advances in neural information processing systems},
  volume={36},
  pages={9156--9172},
  year={2023}
}

@article{hafner2025mastering,
  title={Mastering diverse control tasks through world models},
  author={Hafner, Danijar and Pasukonis, Jurgis and Ba, Jimmy and Lillicrap, Timothy},
  journal={Nature},
  volume={640},
  number={8059},
  pages={647--653},
  year={2025},
  publisher={Nature Publishing Group UK London}
}

@article{team2025evaluating,
  title={Evaluating gemini robotics policies in a veo world simulator},
  author={Team, Gemini Robotics and Choromanski, Krzysztof and Devin, Coline and Du, Yilun and Dwibedi, Debidatta and Gao, Ruiqi and Jindal, Abhishek and Kipf, Thomas and Kirmani, Sean and Leal, Isabel and others},
  journal={arXiv preprint arXiv:2512.10675},
  year={2025}
}

@article{hafner2025training,
  title={Training agents inside of scalable world models},
  author={Hafner, Danijar and Yan, Wilson and Lillicrap, Timothy},
  journal={arXiv preprint arXiv:2509.24527},
  year={2025}
}

@article{assran2025v,
  title={V-jepa 2: Self-supervised video models enable understanding, prediction and planning},
  author={Assran, Mido and Bardes, Adrien and Fan, David and Garrido, Quentin and Howes, Russell and Muckley, Matthew and Rizvi, Ammar and Roberts, Claire and Sinha, Koustuv and Zholus, Artem and others},
  journal={arXiv preprint arXiv:2506.09985},
  year={2025}
}

@article{chen2025large,
  title={Large video planner enables generalizable robot control},
  author={Chen, Boyuan and Zhang, Tianyuan and Geng, Haoran and Zhang, Caiyi and Li, Peihao and Song, Kiwhan and Freeman, William T and Malik, Jitendra and Abbeel, Pieter and Tedrake, Russ and others},
  journal={arXiv preprint arXiv:2512.15840},
  year={2025}
}

@article{jiang2025enerverse,
  title={Enerverse-ac: Envisioning embodied environments with action condition},
  author={Jiang, Yuxin and Chen, Shengcong and Huang, Siyuan and Chen, Liliang and Zhou, Pengfei and Liao, Yue and He, Xindong and Liu, Chiming and Li, Hongsheng and Yao, Maoqing and others},
  journal={arXiv preprint arXiv:2505.09723},
  year={2025}
}

@article{cheang2024gr,
  title={Gr-2: A generative video-language-action model with web-scale knowledge for robot manipulation},
  author={Cheang, Chi-Lam and Chen, Guangzeng and Jing, Ya and Kong, Tao and Li, Hang and Li, Yifeng and Liu, Yuxiao and Wu, Hongtao and Xu, Jiafeng and Yang, Yichu and others},
  journal={arXiv preprint arXiv:2410.06158},
  year={2024}
}

@article{jain2026weaver,
  title={WEAVER, Better, Faster, Longer: An Effective World Model for Robotic Manipulation},
  author={Jain, Arnav Kumar and Wu, Yilin and Farebrother, Jesse and Swamy, Gokul and Bajcsy, Andrea},
  journal={arXiv e-prints},
  pages={arXiv--2606},
  year={2026}
}

@article{yin2026playworld,
  title={Playworld: Learning robot world models from autonomous play},
  author={Yin, Tenny and Mei, Zhiting and Zheng, Zhonghe and Yamane, Miyu and Wang, David and Sceats, Jade and Bateman, Samuel M and Zha, Lihan and Badithela, Apurva and Shorinwa, Ola and others},
  journal={arXiv preprint arXiv:2603.09030},
  year={2026}
}

@inproceedings{
    li2026dworldeval,
    title={Scaling Robot Policy Evaluation via Discrete Diffusion World Model},
    author={Yaxuan Li and Junjie Wen and Zhongyi Zhou and Yefei Chen and Chaomin Shen and Yaxin Peng and Yichen Zhu},
    booktitle={Forty-third International Conference on Machine Learning},
    year={2026},
    url={https://openreview.net/forum?id=93uNlQ1Qp0}
}

@article{agarwal2026cosmos,
  title={Cosmos 3: Omnimodal world models for physical ai},
  author={Agarwal, Niket and Ali, Arslan and Allen, Jon and Antolini, Martin and Aubame, Adeline and Azzolini, Alisson and Bai, Junjie and Bala, Maciej and Balaji, Yogesh and Bapst, Josh and others},
  journal={arXiv preprint arXiv:2606.02800},
  year={2026}
}

@inproceedings{whitney2024learning,
  title={Learning 3d particle-based simulators from rgb-d videos},
  author={Whitney, William and Lopez-Guevara, Tatiana and Pfaff, Tobias and Rubanova, Yulia and Kipf, Thomas and Stachenfeld, Kimberly and Allen, Kelsey},
  booktitle={International Conference on Learning Representations},
  volume={2024},
  pages={50009--50034},
  year={2024}
}

@article{zhang2024dynamic,
  title={Dynamic 3d gaussian tracking for graph-based neural dynamics modeling},
  author={Zhang, Mingtong and Zhang, Kaifeng and Li, Yunzhu},
  journal={arXiv preprint arXiv:2410.18912},
  year={2024}
}

@article{huang2026soma,
  title={SoMA: A Real-to-Sim Neural Simulator for Robotic Soft-body Manipulation},
  author={Huang, Mu and Wang, Hui and Ren, Kerui and Xu, Linning and Zhou, Yunsong and Yu, Mulin and Dai, Bo and Pang, Jiangmiao},
  journal={arXiv preprint arXiv:2602.02402},
  year={2026}
}

@article{zhang2024adaptigraph,
  title={Adaptigraph: Material-adaptive graph-based neural dynamics for robotic manipulation},
  author={Zhang, Kaifeng and Li, Baoyu and Hauser, Kris and Li, Yunzhu},
  journal={arXiv preprint arXiv:2407.07889},
  year={2024}
}

@inproceedings{driess2023learning,
  title={Learning multi-object dynamics with compositional neural radiance fields},
  author={Driess, Danny and Huang, Zhiao and Li, Yunzhu and Tedrake, Russ and Toussaint, Marc},
  booktitle={Conference on robot learning},
  pages={1755--1768},
  year={2023},
  organization={PMLR}
}

@inproceedings{bauer2024doughnet,
  title={Doughnet: A visual predictive model for topological manipulation of deformable objects},
  author={Bauer, Dominik and Xu, Zhenjia and Song, Shuran},
  booktitle={European Conference on Computer Vision},
  pages={92--108},
  year={2024},
  organization={Springer}
}

@inproceedings{lin2022learning,
  title={Learning visible connectivity dynamics for cloth smoothing},
  author={Lin, Xingyu and Wang, Yufei and Huang, Zixuan and Held, David},
  booktitle={Conference on Robot Learning},
  pages={256--266},
  year={2022},
  organization={PMLR}
}

@article{shi2023robocook,
  title={Robocook: Long-horizon elasto-plastic object manipulation with diverse tools},
  author={Shi, Haochen and Xu, Huazhe and Clarke, Samuel and Li, Yunzhu and Wu, Jiajun},
  journal={arXiv preprint arXiv:2306.14447},
  year={2023}
}

@inproceedings{abou2024physically,
  title={Physically embodied gaussian splatting: A visually learnt and physically grounded 3d representation for robotics},
  author={Abou-Chakra, Jad and Rana, Krishan and Dayoub, Feras and Suenderhauf, Niko},
  booktitle={8th Annual Conference on Robot Learning},
  year={2024}
}

@inproceedings{xie2024physgaussian,
  title={Physgaussian: Physics-integrated 3d gaussians for generative dynamics},
  author={Xie, Tianyi and Zong, Zeshun and Qiu, Yuxing and Li, Xuan and Feng, Yutao and Yang, Yin and Jiang, Chenfanfu},
  booktitle={2024 IEEE/CVF Conference on Computer Vision and Pattern Recognition (CVPR)},
  pages={4389--4398},
  year={2024},
  organization={IEEE}
}

@inproceedings{zhang2024physdreamer,
  title={Physdreamer: Physics-based interaction with 3d objects via video generation},
  author={Zhang, Tianyuan and Yu, Hong-Xing and Wu, Rundi and Feng, Brandon Y and Zheng, Changxi and Snavely, Noah and Wu, Jiajun and Freeman, William T},
  booktitle={European Conference on Computer Vision},
  pages={388--406},
  year={2024},
  organization={Springer}
}

@inproceedings{shao2025gaussim,
  title={Gaussim: Foreseeing reality by gaussian simulator for elastic objects},
  author={Shao, Yidi and Huang, Mu and Loy, Chen Change and Dai, Bo},
  booktitle={2025 IEEE/CVF International Conference on Computer Vision (ICCV)},
  pages={7841--7850},
  year={2025},
  organization={IEEE}
}

@article{li2018learning,
  title={Learning particle dynamics for manipulating rigid bodies, deformable objects, and fluids},
  author={Li, Yunzhu and Wu, Jiajun and Tedrake, Russ and Tenenbaum, Joshua B and Torralba, Antonio},
  journal={arXiv preprint arXiv:1810.01566},
  year={2018}
}

@inproceedings{sanchez2020learning,
  title={Learning to simulate complex physics with graph networks},
  author={Sanchez-Gonzalez, Alvaro and Godwin, Jonathan and Pfaff, Tobias and Ying, Rex and Leskovec, Jure and Battaglia, Peter},
  booktitle={International conference on machine learning},
  pages={8459--8468},
  year={2020},
  organization={Pmlr}
}

@article{pfaff2020learning,
  title={Learning mesh-based simulation with graph networks},
  author={Pfaff, Tobias and Fortunato, Meire and Sanchez-Gonzalez, Alvaro and Battaglia, Peter W},
  journal={arXiv preprint arXiv:2010.03409},
  year={2020}
}

@inproceedings{jiang2025phystwin,
  title={Phystwin: Physics-informed reconstruction and simulation of deformable objects from videos},
  author={Jiang, Hanxiao and Hsu, Hao-Yu and Zhang, Kaifeng and Yu, Hsin-Ni and Wang, Shenlong and Li, Yunzhu},
  booktitle={2025 IEEE/CVF International Conference on Computer Vision (ICCV)},
  pages={7219--7230},
  year={2025},
  organization={IEEE}
}

@article{shi2024robocraft,
  title={Robocraft: Learning to see, simulate, and shape elasto-plastic objects in 3d with graph networks},
  author={Shi, Haochen and Xu, Huazhe and Huang, Zhiao and Li, Yunzhu and Wu, Jiajun},
  journal={The International Journal of Robotics Research},
  volume={43},
  number={4},
  pages={533--549},
  year={2024},
  publisher={SAGE Publications Sage UK: London, England}
}

@inproceedings{fan2017point,
  title={A Point Set Generation Network for 3D Object Reconstruction from a Single Image},
  author={Fan, Haoqiang and Su, Hao and Guibas, Leonidas J.},
  booktitle={Proceedings of the IEEE Conference on Computer Vision and Pattern Recognition (CVPR)},
  year={2017}
}

@inproceedings{cicek20163d,
  title={3D U-Net: Learning Dense Volumetric Segmentation from Sparse Annotation},
  author={{\c{C}}i{\c{c}}ek, {\"O}zg{\"u}n and Abdulkadir, Ahmed and Lienkamp, Soeren S. and Brox, Thomas and Ronneberger, Olaf},
  booktitle={Medical Image Computing and Computer-Assisted Intervention (MICCAI)},
  pages={424--432},
  year={2016}
}

@inproceedings{vaswani2017attention,
  title={Attention Is All You Need},
  author={Vaswani, Ashish and Shazeer, Noam and Parmar, Niki and Uszkoreit, Jakob and Jones, Llion and Gomez, Aidan N. and Kaiser, Lukasz and Polosukhin, Illia},
  booktitle={Advances in Neural Information Processing Systems},
  volume={30},
  pages={5998--6008},
  year={2017}
}

@inproceedings{bertasius2021space,
  title={Is Space-Time Attention All You Need for Video Understanding?},
  author={Bertasius, Gedas and Wang, Heng and Torresani, Lorenzo},
  booktitle={Proceedings of the 38th International Conference on Machine Learning},
  pages={813--824},
  year={2021}
}

@inproceedings{bengio2015scheduled,
  title={Scheduled Sampling for Sequence Prediction with Recurrent Neural Networks},
  author={Bengio, Samy and Vinyals, Oriol and Jaitly, Navdeep and Shazeer, Noam},
  booktitle={Advances in Neural Information Processing Systems},
  volume={28},
  pages={1171--1179},
  year={2015}
}

@article{zhang2026contactworld,
  title={ContactWorld: What Matters in Vision-Tactile World Models for Contact-Rich Manipulation},
  author={Zhang, Zhiyuan and Zhou, Pokuang and Zhang, Kaidi and Desai, Adeesh and Amosa, Temitope and Soleymanzadeh, Davood and Lei, Jiuzhou and Zheng, Minghui and She, Yu},
  journal={arXiv preprint arXiv:2606.13877},
  year={2026}
}

@article{huang2026vitacworld,
  title={ViTacWorld: Scaling visuo-tactile world models for contact-rich robot manipulation},
  author={Huang, Yunao and Sang, Shiyu and Lu, Haotao and Ni, Suting and Wu, Shijie and Guo, Ziyang and Shi, Ye and Wang, Jingya},
  journal={arXiv preprint arXiv:2607.22530},
  year={2026}
}

@article{ma2026feelworld,
  title={FeelWorld: Visuo-Tactile World Model for Hierarchical Contact Prediction and Planning},
  author={Ma, Wenxuan and Zhang, Chaofan and Xue, Chao and Cai, Yinghao and Yao, Guocai and Cui, Shaowei and Wang, Shuo},
  journal={arXiv preprint arXiv:2607.24267},
  year={2026}
}

@article{ai2024robopack,
  title={Robopack: Learning tactile-informed dynamics models for dense packing},
  author={Ai, Bo and Tian, Stephen and Shi, Haochen and Wang, Yixuan and Tan, Cheston and Li, Yunzhu and Wu, Jiajun},
  journal={arXiv preprint arXiv:2407.01418},
  year={2024}
}

@article{wu2026inference,
  title={Inference-time Policy Steering via Vision and Touch},
  author={Wu, Yilin and Si, Zilin and Temel, Zeynep and Kroemer, Oliver and Bajcsy, Andrea},
  journal={arXiv preprint arXiv:2606.14981},
  year={2026}
}

@inproceedings{tian2019manipulation,
  title={Manipulation by feel: Touch-based control with deep predictive models},
  author={Tian, Stephen and Ebert, Frederik and Jayaraman, Dinesh and Mudigonda, Mayur and Finn, Chelsea and Calandra, Roberto and Levine, Sergey},
  booktitle={2019 International Conference on Robotics and Automation (ICRA)},
  pages={818--824},
  year={2019},
  organization={IEEE}
}

@article{lou2026dream,
  title={Dream-Tac: A Unified Tactile World Action Model for Contact-Rich Robot Manipulation},
  author={Lou, Yunfan and Ye, Yifan and Fu, Yankai and Cen, Jun and Chi, Xiaowei and Lyu, Yaoxu and Jia, Peidong and Han, Sirui and Lu, Zhihe and Zhang, Shanghang},
  journal={arXiv preprint arXiv:2606.08737},
  year={2026}
}

@article{ye2025learning,
  title={Learning to feel the future: Dreamtacvla for contact-rich manipulation},
  author={Ye, Guo and Zhang, Zexi and Zhao, Xu and Wu, Shang and Lu, Haoran and Lu, Shihan and Liu, Han},
  journal={arXiv preprint arXiv:2512.23864},
  year={2025}
}

@article{zang2026tacforesight,
  title={TacForeSight: Force-Guided Tactile World Model for Contact-Rich Manipulation},
  author={Zang, Yujie and Zheng, Yuhang and Nie, Xian and Zheng, Yupeng and Tian, Shuai and Gu, Songen and Gao, Chen and Wang, Zining and Yan, Shuicheng and Ding, Wenchao},
  journal={arXiv preprint arXiv:2606.11184},
  year={2026}
}

@article{yuan2026vtam,
  title={Vtam: Video-tactile-action models for complex physical interaction beyond vlas},
  author={Yuan, Haoran and Yi, Weigang and Zhang, Zhenyu and Chen, Wendi and Mo, Yuchen and Yin, Jiashi and Li, Xinzhuo and Zeng, Xiangyu and Wen, Chuan and Lu, Cewu and others},
  journal={arXiv preprint arXiv:2603.23481},
  year={2026}
}

@article{wu2026tactile,
  title={Tactile-wam: Touch-aware world action model with tactile asymmetric attention},
  author={Wu, Siyu and You, Linjing and Zhu, Junjie and Liu, Yaozu and Kaixiang, Huang and Yonghang, Chen and Li, Jituo and Zhang, Changhao and Liu, Jian and Hengshuo, Zhu and others},
  journal={arXiv preprint arXiv:2606.26663},
  year={2026}
}

@inproceedings{eigen2014depth,
  title={Depth Map Prediction from a Single Image using a Multi-Scale Deep Network},
  author={Eigen, David and Puhrsch, Christian and Fergus, Rob},
  booktitle={Advances in Neural Information Processing Systems},
  volume={27},
  pages={2366--2374},
  year={2014}
}

@inproceedings{kim2026cosmos,
  title={Cosmos Policy: Fine-Tuning Video Models for Visuomotor Control and Planning},
  author={Kim, Moo Jin and Gao, Yihuai and Lin, Tsung-Yi and Lin, Yen-Chen and Ge, Yunhao and Lam, Grace and Liang, Percy and Song, Shuran and Liu, Ming-Yu and Finn, Chelsea and Gu, Jinwei},
  booktitle={International Conference on Learning Representations},
  year={2026}
}

@article{williams2017model,
  title={Model Predictive Path Integral Control: From Theory to Parallel Computation},
  author={Williams, Grady and Aldrich, Andrew and Theodorou, Evangelos A.},
  journal={Journal of Guidance, Control, and Dynamics},
  volume={40},
  number={2},
  pages={344--357},
  year={2017},
  doi={10.2514/1.G001921}
}

@inproceedings{karaev2024cotracker,
  title={CoTracker: It Is Better to Track Together},
  author={Karaev, Nikita and Rocco, Ignacio and Graham, Benjamin and Neverova, Natalia and Vedaldi, Andrea and Rupprecht, Christian},
  booktitle={European Conference on Computer Vision (ECCV)},
  pages={18--35},
  year={2024}
}

@inproceedings{zhou2019continuity,
  title={On the Continuity of Rotation Representations in Neural Networks},
  author={Zhou, Yi and Barnes, Connelly and Lu, Jingwan and Yang, Jimei and Li, Hao},
  booktitle={Proceedings of the IEEE/CVF Conference on Computer Vision and Pattern Recognition (CVPR)},
  pages={5745--5753},
  year={2019}
}

@inproceedings{graham2018submanifold,
  title={3D Semantic Segmentation with Submanifold Sparse Convolutional Networks},
  author={Graham, Benjamin and Engelcke, Martin and van der Maaten, Laurens},
  booktitle={Proceedings of the IEEE Conference on Computer Vision and Pattern Recognition (CVPR)},
  pages={9224--9232},
  year={2018}
}

@inproceedings{he2016deep,
  title={Deep Residual Learning for Image Recognition},
  author={He, Kaiming and Zhang, Xiangyu and Ren, Shaoqing and Sun, Jian},
  booktitle={Proceedings of the IEEE Conference on Computer Vision and Pattern Recognition (CVPR)},
  pages={770--778},
  year={2016}
}

@inproceedings{hu2022lora,
  title={LoRA: Low-Rank Adaptation of Large Language Models},
  author={Hu, Edward J. and Shen, Yelong and Wallis, Phillip and Allen-Zhu, Zeyuan and Li, Yuanzhi and Wang, Shean and Wang, Lu and Chen, Weizhu},
  booktitle={International Conference on Learning Representations},
  year={2022}
}

@article{huang20243d,
  title={3d-vitac: Learning fine-grained manipulation with visuo-tactile sensing},
  author={Huang, Binghao and Wang, Yixuan and Yang, Xinyi and Luo, Yiyue and Li, Yunzhu},
  journal={arXiv preprint arXiv:2410.24091},
  year={2024}
}

@article{huang2025vt,
  title={Vt-refine: Learning bimanual assembly with visuo-tactile feedback via simulation fine-tuning},
  author={Huang, Binghao and Xu, Jie and Akinola, Iretiayo and Yang, Wei and Sundaralingam, Balakumar and O'Flaherty, Rowland and Fox, Dieter and Wang, Xiaolong and Mousavian, Arsalan and Chao, Yu-Wei and others},
  journal={arXiv preprint arXiv:2510.14930},
  year={2025}
}
\bibliographystyle{iclr2027_conference}

\appendix

\section*{Appendix Contents}

\noindent
\textbf{A. Further Experimental Results} \dotfill \pageref{appendix:policy-training-details}

\hspace{1em}
A.1 Deform360 Results \dotfill \pageref{appendix:deform360-results}

\noindent
\textbf{B. Model and Training Details} \dotfill \pageref{appendix:model-details}

\hspace{1em}
B.1 Hybrid Particle-Voxel Autoencoder Details \dotfill \pageref{appendix:vae}

\hspace{1em}
B.2 3D Latent Dynamics Model Details \dotfill \pageref{appendix:3d-latent-dynamics}

\noindent
\textbf{C. ALOHA Task Details} \dotfill \pageref{appendix:aloha-task-setup}

\hspace{1em}
C.1 ALOHA Task Description and Setup \dotfill \pageref{appendix: aloha-task-description-and-setup}

\hspace{1em}
C.2 Qualitative Visualizations of Open-Loop Rollouts\dotfill \pageref{appendix: aloha-task-description-and-setup}

\hspace{1em}
C.3 Baseline Details\dotfill \pageref{appendix:baselines}

\noindent
\textbf{D. Policy Training and Evaluation Details} \dotfill \pageref{appendix:policy-training}

\hspace{1em}
D.1 Evaluation Criteria\dotfill \pageref{appendix:evaluation-criteria}

\hspace{1em}
D.2 Task Details\dotfill \pageref{appendix:task-details}

\hspace{1em}
D.3 Policy Training Details\dotfill \pageref{appendix:policy-training-details}

\hspace{1em}
D.4 Visualization of Real Demonstrations\dotfill \pageref{appendix:visualization-real-demo}

\hspace{1em}
D.5 Visualization of World Model Policy Input\dotfill \pageref{appendix:visualization-policy}

\section{Further Experimental Results}
\label{appendix:further-experimental-results}

In this section, we provide further experimental details for our method. 

\subsubsection{Deform360 Results}
\label{appendix:deform360-results}

\begin{center}
\small
\resizebox{0.8\textwidth}{!}{%
\begin{tabular}{@{}c@{\hspace{12pt}}lcccccc@{}}
\toprule
& & \multicolumn{3}{c}{Training objects} & \multicolumn{3}{c}{Held-out objects} \\
\cmidrule(lr){3-5} \cmidrule(lr){6-8}
& Method & PSNR $\uparrow$ & LPIPS $\downarrow$ & CD (mm) $\downarrow$
& PSNR $\uparrow$ & LPIPS $\downarrow$ & CD (mm) $\downarrow$ \\
\midrule
\multirow{3}{*}{\rotatebox{90}{\scriptsize \textbf{Particle}}}
& PGND & 14.9 & 0.41 & 18.6 & 12.3 & 0.50 & 25.2 \\
& PointWorld & 16.8 & 0.34 & 12.4 & 15.0 & 0.58 & 14.5 \\
& ParticleFormer & 18.0 & 0.30 & 10.5 & \textbf{15.6} & \textbf{0.31} & 13.4 \\
\midrule
& \textbf{Ours} & \textbf{18.6} & \textbf{0.26} & \textbf{9.6}
& 14.8 & 0.43 & \textbf{12.1} \\
\bottomrule
\end{tabular}%
}

\captionof{table}{\small \textbf{Quantitative Comparison on Deform360.}
Results are averaged over 150 step open-loop rollouts. On objects seen during training, 3D World Simulator outperforms all methods on both image and geometry metrics. On held-out objects, our method retains better geometry but worse rendering quality. This advantage can be explained by particle methods outputting tracks on fitted Gaussian Splats, while our method generates appearance on unseen objects. }
\label{tab:quantitative-deform360}
\end{center}
\section{Model and Training Details}
\label{appendix:model-details}

We describe the autoencoder and dynamics model details for both ALOHA datasets and the Deform360 dataset. 

\subsection{Hybrid Particle--Voxel Autoencoder}
\label{appendix:vae}

\paragraph{Representation.}
ALOHA observations represent the cropped manipulation workspace,
including robot, object, and tactile observations when available.
Deform360 observations represent the object point cloud.
All positions are expressed in a common coordinate frame before
voxelization.

The encoder produces an eight-channel latent grid for both ALOHA and Deform360 datasets.

\paragraph{Decoder.}
The voxel decoder predicts occupancy at three resolutions and
color at the finest resolution.
Occupied voxels seed two successive particle-refinement stages.
The particle head predicts continuous positions and Gaussian
rendering attributes, mainly anisotropic Gaussians and
degree-one spherical harmonics.

\paragraph{Voxel supervision.}
We supervise occupancy using binary
cross-entropy at each decoder resolution.
Voxel RGB is supervised with an $\ell_1$ loss at occupied cells, which includes tactile particles and visual particles.

Thus,
\begin{equation}
\mathcal{L}_{\mathrm{voxel}}
=
\sum_{\ell}\lambda_{\mathrm{occ},\ell}
\mathcal{L}_{\mathrm{BCE}}^{(\ell)}
\end{equation}

For tactile observations, we additionally supervise tactile
occupancy with binary cross-entropy and tactile signal values
with an $\ell_1$ loss at tactile-occupied cells. The tactile terms are omitted when tactile supervision
is unavailable. Spatial weighting emphasizes object and
contact regions.

\begin{table}[t]
\centering
\small
\caption{Exact autoencoder configurations for each dataset.}
\label{tab:ae-settings}
\begin{tabular}{@{}lcc@{}}
\toprule
Setting & ALOHA & Deform360 \\
\midrule
Latent grid $(X,Y,Z)$
& $32 \times 16 \times 32$
& $32 \times 32 \times 32$ \\
Latent channels 
& $8$ & $8$ \\
Workspace extent $(X,Y,Z)$, metres
& $0.52 \times 0.32 \times 0.28$
& $0.515 \times 0.525 \times 0.38$ \\
Sparse encoder widths
& $8,16,32,64$ & $8,16,32,64$ \\
Voxel decoder widths
& $256,128,64$ & $320,160,80$ \\
Particle-head hidden width & $64$ & $64$ \\
Particle refinement stages & $2$ & $2$ \\
Gaussian scales & Anisotropic & Anisotropic \\
Spherical-harmonic degree & $1$ & $1$ \\
Tactile supervision & Enabled & Disabled \\
Maximum input points & $300{,}000$ & $50{,}000$ \\
Batch size per process & $1$ & $1$ \\
Initial learning rate
& $2 \times 10^{-4}$ & $2 \times 10^{-4}$ \\
Minimum learning rate & $10^{-5}$ & $10^{-5}$ \\
Weight decay & $10^{-4}$ & $10^{-4}$ \\
Latent noise standard deviation & $0.1$ & $0.1$ \\
Consistency coefficient & $100$ & $100$ \\
KL regularization & Disabled & Disabled \\
\bottomrule
\end{tabular}
\end{table}

\paragraph{Particle supervision.}
For each occupied seed voxel, let $\hat{P}$ denote its
predicted particles and $P$ the ground-truth points in its
local voxel neighborhood. We use the local bidirectional
Chamfer distance
\begin{equation}
\label{eq: chamfer}
\mathcal{L}_{\mathrm{CD}}
=
\frac{1}{|\hat{P}|}
\sum_{\hat{p}\in\hat{P}}
\min_{p\in P}\|\hat{p}-p\|_2
+
\frac{1}{|P|}
\sum_{p\in P}
\min_{\hat{p}\in\hat{P}}\|p-\hat{p}\|_2.
\end{equation}
We average this objective over valid seed neighborhoods and
apply it at both particle-refinement stages.
Neighborhoods without ground-truth support are excluded from
this local matching term.

For color supervision, ground-truth points are assigned to
their nearest predicted particle within the neighborhood.
The mean color of the assigned points serves as an
$\ell_1$ target for that particle.
The combined particle objective is
\begin{equation}
\mathcal{L}_{\mathrm{particle}}
=
\sum_r
\left(
\mathcal{L}_{\mathrm{CD}}^{(r)}
+
\mathcal{L}_{\mathrm{rgb}}^{(r)}
\right).
\end{equation}

\paragraph{Rendering supervision.}
For a training camera $j$, the predicted Gaussians produce
an RGB image $\hat{I}_j$, depth $\hat{D}_j$, and opacity.
Rendering supervision is
\begin{equation}
\begin{split}
\mathcal{L}_{\mathrm{render}}
={}&
\lambda_{\mathrm{RGB}}
\left(
\mathcal{L}_{1}^{\mathrm{RGB}}
+
\lambda_{\mathrm{SSIM}}\mathcal{L}_{\mathrm{SSIM}}
+
\mathcal{L}_{\mathrm{LPIPS}}
\right) + \mathcal{L}_{1}^{\mathrm{Depth}}
\end{split}
\end{equation}
where
\begin{equation}
\mathcal{L}_{\mathrm{SSIM}}
=
\frac{1-\operatorname{SSIM}(\hat{I}_j,I_j)}{2},
\end{equation}
The image losses are aggregated over supervised camera views.
We use $\lambda_{\mathrm{RGB}} = 4$, $\lambda_{\mathrm{SSIM}} = 0.5$.

\paragraph{Subsampling consistency.}
Two independently sampled point clouds from the same observation
are encoded with shared weights. We minimize
\begin{equation}
\mathcal{L}_{\mathrm{cons}}
=
\frac{1}{c|\Omega_t|}
\sum_{v\in\Omega_t}
\left\|
E_\theta(P_t^{(1)})(v)
-
E_\theta(P_t^{(2)})(v)
\right\|_2^2,
\end{equation}
where $\Omega_t$ contains occupied latent cells.
When color augmentation is used, both subsamples receive the
same color transformation, so consistency does not encourage
invariance to object appearance.

\paragraph{Training stages.}
The reference recipe first trains the voxel autoencoder for
10 epochs. During the following 10 epochs, the voxel
autoencoder is frozen while the particle and rendering heads
are trained. Subsequent training jointly optimizes the
autoencoder and particle heads.

\subsection{3D Latent Dynamics}
\label{appendix:3d-latent-dynamics}

\paragraph{Per-frame spatial encoder.}
Let the observed history contain $H$ latent grids, including
the most recent observation:
$z_{t-H+1},\ldots,z_t$.
A shared 3D convolutional encoder processes each frame
independently, preserving an explicit temporal dimension.
Two spatial downsampling stages produce features at
one-half and one-quarter of the latent-grid resolution.

\paragraph{Action conditioning.}
Each observed frame is paired with its corresponding
robot state or action descriptor.
A future query slot is conditioned on the action for the
requested prediction.
An MLP embeds the action, relative timestamp, and history-noise
level, and injects this conditioning into the feature blocks.

For Deform360, the action descriptor contains each gripper's
3D position, 6D orientation representation~\citep{zhou2019continuity}, opening, and
presence indicator.
Positions and orientations are expressed in the same frame
as the object latent.
For ALOHA, the corresponding descriptor is obtained from the
robot kinematics and gripper state.

Optionally, we concatenate four spatial action channels to
each frame: a position heatmap and an opening-modulated
heatmap for each gripper.
The future query receives the target action's spatial channels.
Missing grippers produce zero-valued spatial channels.

\paragraph{Factorized spatial--temporal attention.}
At the coarsest resolution, the model alternates spatial
and temporal attention.
Spatial attention operates over all coarse voxels within
each frame:
\begin{equation}
\bar{h}_{\tau}
=
h_{\tau}
+
\operatorname{SA}\!\left(
\operatorname{LN}(h_{\tau})
\right).
\end{equation}
Temporal attention then operates across frame slots at
each coarse spatial location:
\begin{equation}
h'_{\tau,v}
=
\bar{h}_{\tau,v}
+
\operatorname{TA}_{\mathrm{causal}}\!\left(
\operatorname{LN}(\bar{h}_{\leq\tau,v})
\right).
\end{equation}
Each attention block includes a residual feed-forward network.
Spatial positions and temporal offsets are encoded explicitly.
The future query may attend to all observed frames but never
to future ground-truth latent features.

For a $32^3$ latent, attention is applied to $8^3=512$
spatial tokens per frame. This avoids attention over all
$32^3$ fine-grid positions.
For the ALOHA grid, the corresponding coarse grid is
$8\times4\times8$.

\paragraph{Decoder and skip connections.}
A shared 3D decoder upsamples the attention features back
to the original latent resolution.
The predicted future slot uses skip features from the
most recent available history frame.
During open-loop prediction, this frame may itself be
a previous model prediction.
No future ground-truth features are used in the skip paths.
The output is a direct latent prediction and requires
one network evaluation per predicted step.

\begin{table}[t]
\centering
\small
\caption{3D latent dynamics configuration.}
\label{tab:proposed-dynamics}
\begin{tabular}{@{}lcc@{}}
\toprule
Setting & ALOHA & Deform360 \\
\midrule
Latent grid & $32\times16\times32$ & $32^3$ \\
Latent channels & $8$ & $8$ \\
History frames & $5$ & $5$ \\
Encoder widths & $128,256,512$ & $128,256,512$ \\
Convolution kernel & $3^3$ & $3^3$ \\
Spatial downsampling stages & $2$ & $2$ \\
Attention grid & $8\times4\times8$ & $8^3$ \\
Spatial--temporal block pairs & $4$ & $4$ \\
Attention heads & $8$ & $8$ \\
Attention head dimension & $64$ & $64$ \\
Feed-forward expansion & $2$ & $2$ \\
Action embedding dimension & $512$ & $512$ \\
Predictions per forward pass & $1$ & $1$ \\
Optimizer & AdamW & AdamW \\
Weight decay & $10^{-4}$ & $10^{-4}$ \\
Gradient clipping & $10$ & $10$ \\
Maximum history-noise std. & $0.4$ & $0.4$ \\
Rollout discount $\gamma$ & $0.95$ & $0.95$ \\
\bottomrule
\end{tabular}
\end{table}

\paragraph{Autoregressive training.}
We precompute latent sequences using the frozen autoencoder.
Each training example begins with observed history.
The model predicts one future latent, appends it to the
history, removes the oldest history frame, and repeats.
Ground-truth future latents are used only as loss targets.

We use $K$ to denote the number of supervised predictions,
so $K=1$ corresponds to next-frame-only training.
This avoids ambiguity with implementations that count only
the additional feedback steps. We use a rollout discount of $\gamma=0.95$ for the
supervised latent predictions, as defined in Section~\ref{sec:dyn}.

For each history frame, we sample
$\sigma_\tau\sim\mathcal{U}(0,\sigma_{\max})$ and use
\begin{equation}
\tilde{z}_\tau
=
z_\tau+\sigma_\tau\epsilon_\tau,
\qquad
\epsilon_\tau\sim\mathcal{N}(0,I).
\end{equation}
The network receives $\sigma_\tau$ as conditioning.

\section{ALOHA Dataset Task Setup}
\label{appendix:aloha-task-setup}
\textbf{Task Descriptions}

\subsection{ALOHA Task Descriptions and Setup}
\label{appendix: aloha-task-description-and-setup}

\textit{Part Assembly}: The left robot arm grasps the socket while the right arm grasps the plug in various starting positions. The task involves reorienting the plug by the right arm pressing down on the socket, then lifting to insert into the socket. 

\textit{Cloth Folding}: The left robot arm starts by grasping the top left edge of the cloth and lifting it such that the top edge of the cloth straightens. The right robot arm slides its hand across this edge to unfurl the cloth. Then both arms are used to fold the cloth.  

\textit{T Pushing}: Both robot grippers are closed and constrained to the table plane. They are used to push the T block into various positions. 

\textit{Marker Insertion}: The left robot arm grasps the marker in the pencil cup. The right robot arm then grasps the marker and performs in-hand adjustment through sliding and reorienting the marker to a 90 degree angle. The left hand then inserts the marker into the holder. 

\textit{Rope Threading}: The rope starts in a curled and overlapping positions. The right robot arm grasps one end of the rope, while the left robot arm repeatedly slides along the rope until it reaches the other end. Once the left robot arm grasps near the other end of the rope, it inserts it into a pad eye. 

\textit{Box Packing}: The lid of the box starts open and laid on the table. Both robot grippers are used to lift the lid of the box and close it. 

\textbf{Camera Setup}

We utilize three RealSense D455 static stereo cameras including one top-down view camera, and two forward facing cameras on either side of the workspace. In addition, we utilize the two wrist-mounted RealSense D405 cameras, one mounted on each arm.

\textbf{Data Processing} 

To obtain the pointcloud observation, we utilize FoundationStereo~\citep{wen2025foundationstereo} to process the stereo observations from each RealSense camera, before back-projecting into the world frame. To obtain tactile-particle locations from the FlexiTac tactile signals, we use forward kinematics using the ALOHA joint positions to calculate the tactile pad positions. We then crop the point cloud using a manually set workspace bounding box. 

\subsection{Qualitative Visualizations of Open-Loop Rollouts}
\label{appendix:qualitative-rollouts}
Below, we visualize open-loop multi-view renders and point clouds for each of the six tasks from the 3D World Simulator.

\begin{center}
    \includegraphics[width=1\linewidth]{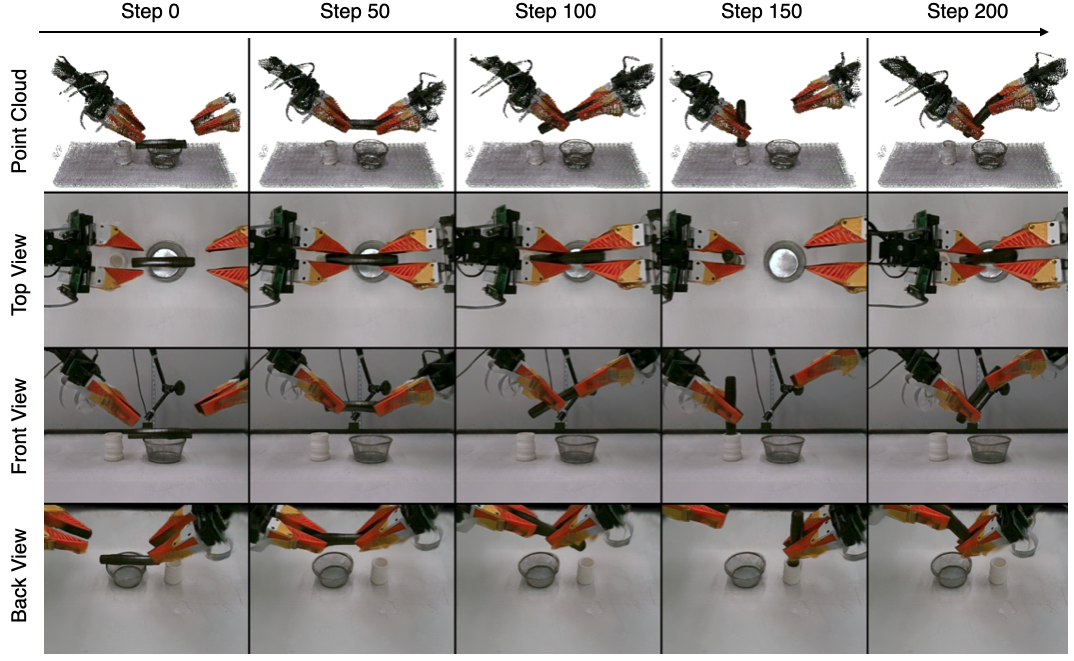}
    \captionof{figure}{\textbf{Marker Insertion open-loop rollout visualization.}}
\end{center}

\begin{center}
    \includegraphics[width=1\linewidth]{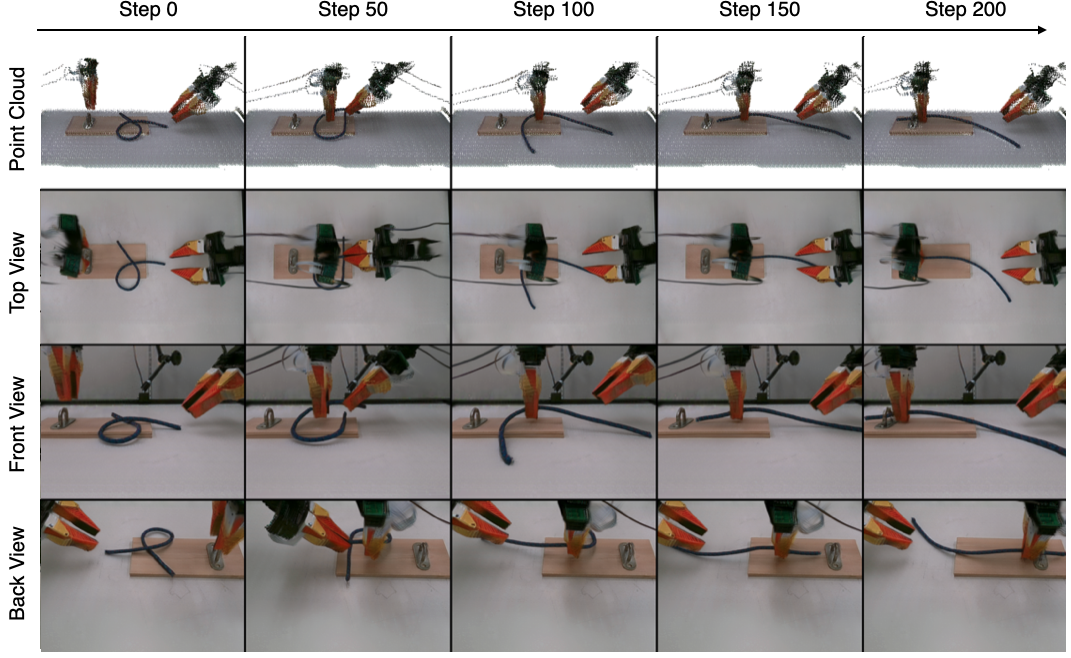}
    \captionof{figure}{\textbf{Rope Threading open-loop rollout visualization.}}
\end{center}

\begin{center}
    \includegraphics[width=1\linewidth]{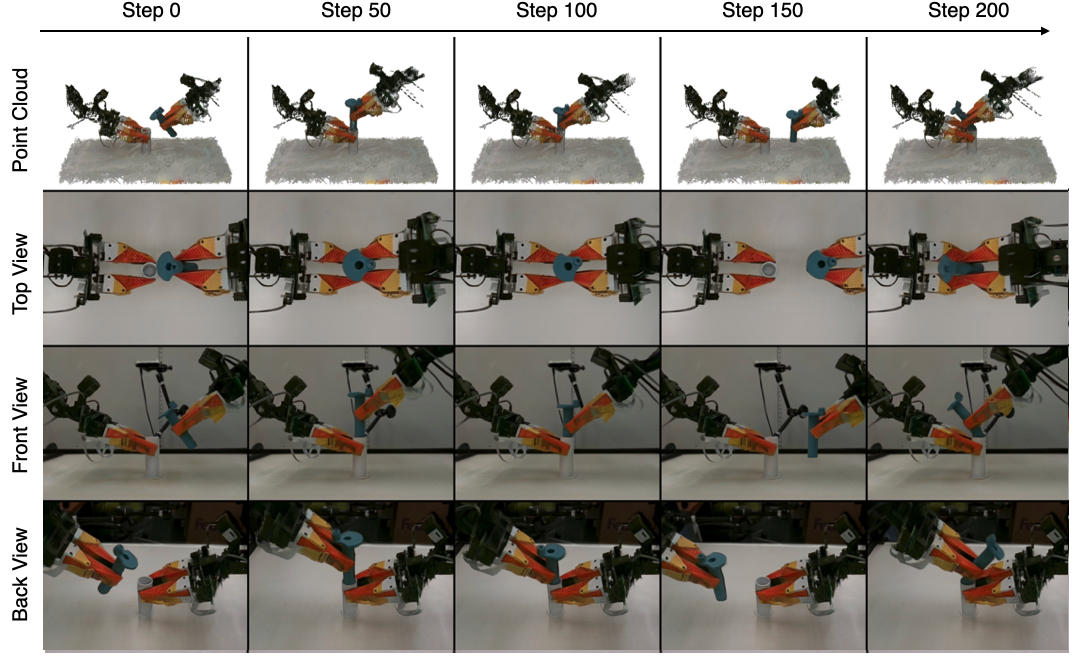}
    \captionof{figure}{\textbf{Part Assembly open-loop rollout visualization.}}
\end{center}

\begin{center}
    \includegraphics[width=1\linewidth]{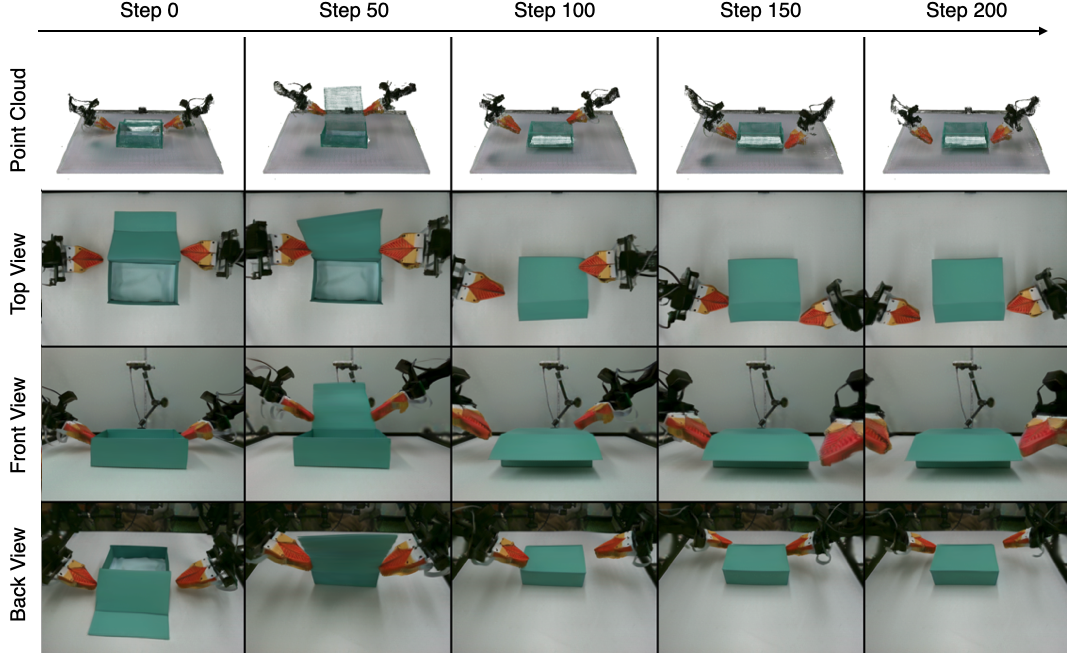}
    \captionof{figure}{\textbf{Box Packing open-loop rollout visualization.}}
\end{center}

\begin{center}
    \includegraphics[width=1\linewidth]{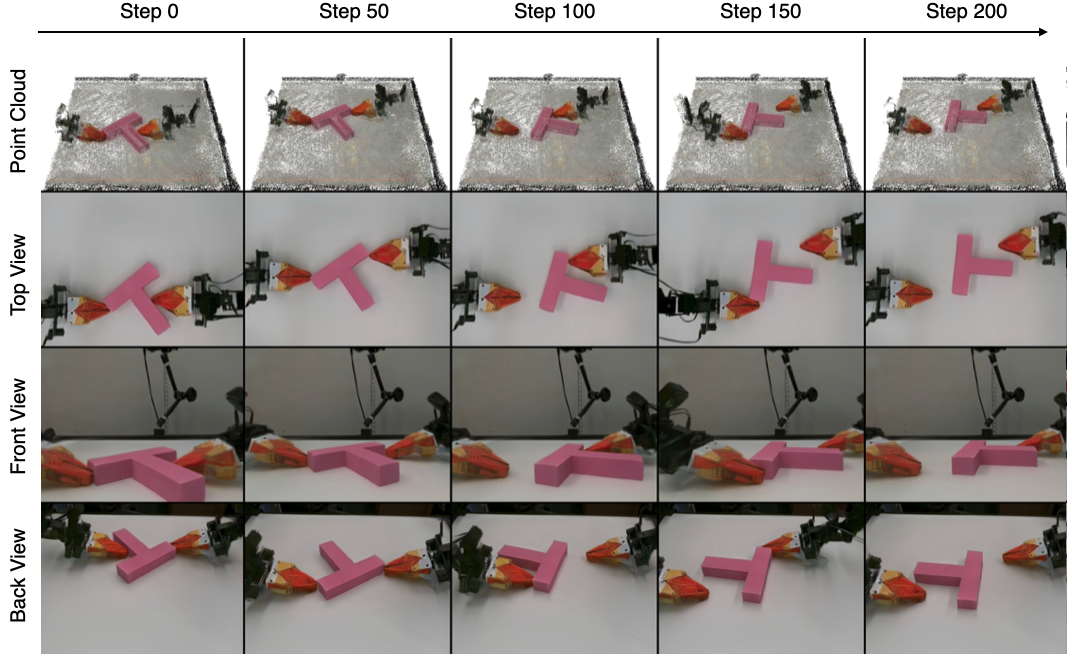}
     \captionof{figure}{\textbf{T Pushing open-loop rollout visualization.}}
\end{center}

\begin{center}
    \includegraphics[width=1\linewidth]{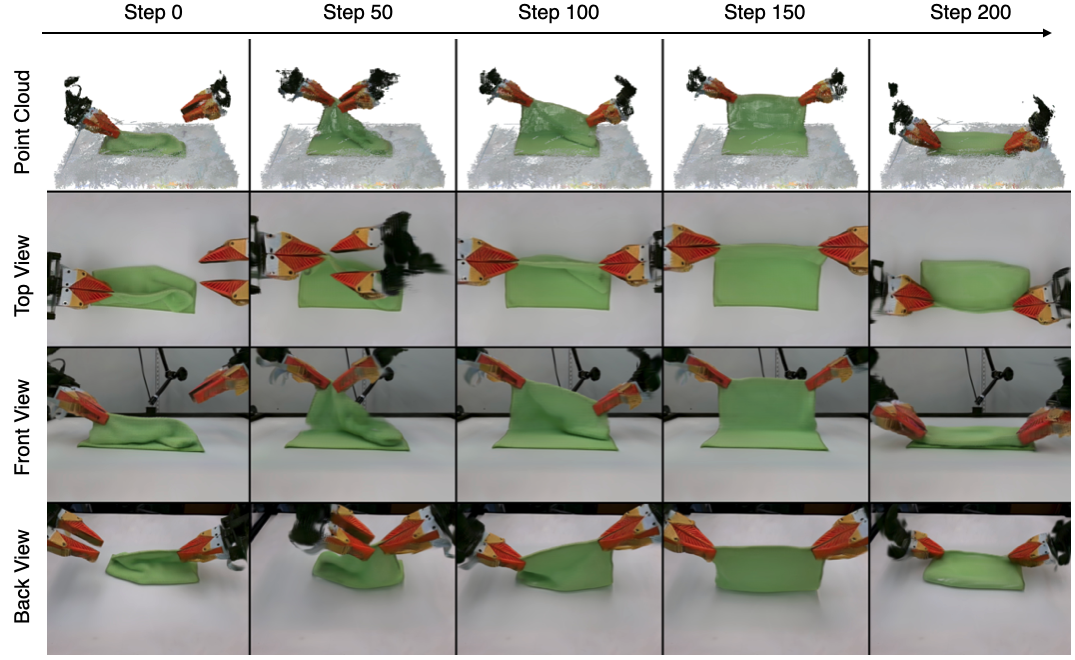}
    \captionof{figure}{\textbf{Cloth Folding open-loop rollout visualization.}}
\end{center}

\subsection{Baseline Details}
\label{appendix:baselines}

To enable Cosmos to predict depth information and allow multi-view renders, we follow guidelines and concatenate four views in one image, in addition to depth mapped into RGB space. For IWS, we enable the model to predict depth in addition to RGB, and concatenate a latent for each view in accordance to training guidelines. We follow the recipe to train Robot4DGen as it is already designed to be 3D aware and enable multi-view rendering.  To train the 3D particle prediction baselines, we utilize CoTracker~\citep{karaev2024cotracker} to obtain point tracks for each object as well as for scene points. 

\section{Policy Training and Evaluation Details}
\label{appendix:policy-training}

\subsection{Evaluation Criteria}
\label{appendix:evaluation-criteria}
For evaluation, we define a success in the Assembly task to be whether the plug is successfully inserted into the socket. For cloth folding, if the edge is not grasped, that is judged to be a failure. If the edge is grasped and the mask IoU is higher than 0.6, we judge that to be a success. For T-Pushing, we define a success to be if the mask IoU is above 0.60. 

\subsection{Task Details}
\label{appendix:task-details}

\paragraph{Part Assembly.} The policy observes only the right wrist-mounted camera, while the socket is heavily occluded in the left hand, making successful insertion dependent on both accurate wrist-camera observations and tactile feedback.

\paragraph{T Pushing.} We utilize a novel viewpoint that was not seen during world model training. Performance is slightly lower most probably because this novel viewpoint is harder than the top down view for successful task performance, rather than the rendering. 

\paragraph{Cloth Folding.} The policy is required to slide along the cloth edge which is not visible from the front camera observation. This task shows the usage of multi-view observations. 

\subsection{Policy Training Details}
\label{appendix:policy-training-details}

We train ACT, Diffusion Policy (DP), $\pi_{0.5}$, and Cosmos on Cloth Folding, Part Assembly, and T-Pushing. Each task contains 100 demonstrations recorded at 10 Hz, with 90 demonstrations used for training and 10 held out. Cloth Folding uses front and wrist-mounted camera observations, Part Assembly uses the wrist-mounted camera together with tactile observations, and T-Pushing uses a single novel camera viewpoint. Real images are color-corrected, cropped, and resized to match the corresponding simulated observations. ACT and DP use $128\times128$ image inputs, while $\pi_{0.5}$ and Cosmos use $224\times224$ inputs.

For Part Assembly, tactile observations consist of four $12\times32$ tactile pads. We apply a small deadband before normalization and use small tactile augmentation during training; Cloth Folding and T-Pushing do not use tactile inputs.

ACT and DP are trained with batch size 64 for 30,000 updates, while $\pi_{0.5}$ uses batch size 8 and is trained for 30,000 updates. ACT predicts 30-action chunks, DP predicts 16-action horizons, and $\pi_{0.5}$ predicts 30-action chunks. We use the optimizer and learning-rate settings listed in Table~\ref{tab:policy-training-details}.

ACT uses an ImageNet-initialized ResNet-18~\citep{he2016deep} visual encoder and transformer policy. DP uses a ResNet-18 visual encoder with a conditional temporal U-Net diffusion model. $\pi_{0.5}$ is fine-tuned using LoRA~\citep{hu2022lora}, while Cosmos is initialized from \texttt{Cosmos-Policy-ALOHA-Predict2-2B}.
We use standard image augmentation techniques like random color jitter, random crop and blur. We also add tactile augmentation to the Part Assembly policy by adding sparse noise, dropout, and Gaussian noise. 

% \begin{table}[h]
% \centering
% \small
% \caption{Policy training configurations.}
% \label{tab:policy-training-details}
% \begin{tabular}{lccc}
% \toprule
% & ACT & DP & $\pi_{0.5}$ \\
% \midrule
% Input resolution & $128^2$ & $128^2$ & $224^2$ \\
% Observation history & 1 & 2 & 1 \\
% Action horizon & 30 & 16 & 30 \\
% Batch size & 64 & 64 & 8 \\
% Learning rate & $10^{-5}$ & $10^{-4}$ & $2.5{\times}10^{-5}$ \\
% Optimizer & AdamW & Adam & AdamW \\
% Training updates & 30k & 30k & 30k \\
% \bottomrule
% \end{tabular}
% \end{table}

\begin{table}[h]
  \centering
  \small
  \caption{Policy training configurations.}
  \label{tab:policy-training-details}
  \begin{tabular}{lcccc}
  \toprule
  & ACT & DP & $\pi_{0.5}$ & Cosmos Policy \\
  \midrule
  Input resolution & $128^2$ & $128^2$ & $224^2$ & $224^2$ \\
  Observation history & 1 & 2 & 1 & 1 \\
  Action horizon & 30 & 16 & 30 & 30 \\
  Batch size & 64 & 64 & 8 & 192 \\
  Learning rate & $10^{-5}$ & $10^{-4}$ & $2.5{\times}10^{-5}$ & $10^{-4}$ \\
  Optimizer & AdamW & Adam & AdamW & AdamW \\
  Training updates & 30k & 30k & 30k & 10k \\
  \bottomrule
  \end{tabular}
  \end{table}

\subsection{Visualization of Real Demos}
\label{appendix:visualization-real-demo}
We visualize collected demonstrations for each task. 

\begin{center}
    \includegraphics[width=1\linewidth]{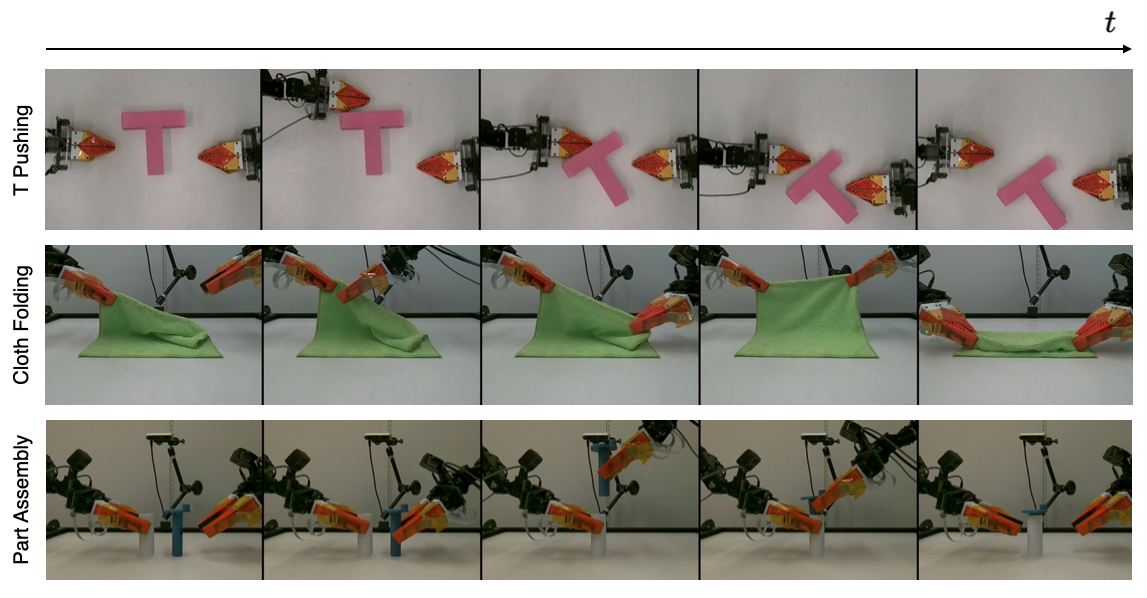}
    \captionof{figure}{\textbf{Visualizations of real world imitation learning data for the cloth folding, part assembly, and T pushing task.}}
\end{center}

\subsection{Qualitative Visualization of World Model Policy Data }
\label{appendix:visualization-policy}

\begin{center}
    \includegraphics[width=1\linewidth]{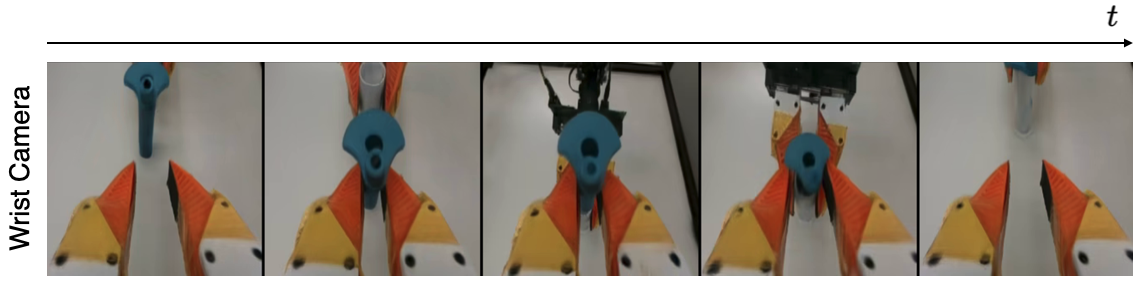}
    \captionof{figure}{\textbf{Visualization of part assembly world model wrist camera input to policies. }}
\end{center}

\begin{center}
    \includegraphics[width=1\linewidth]{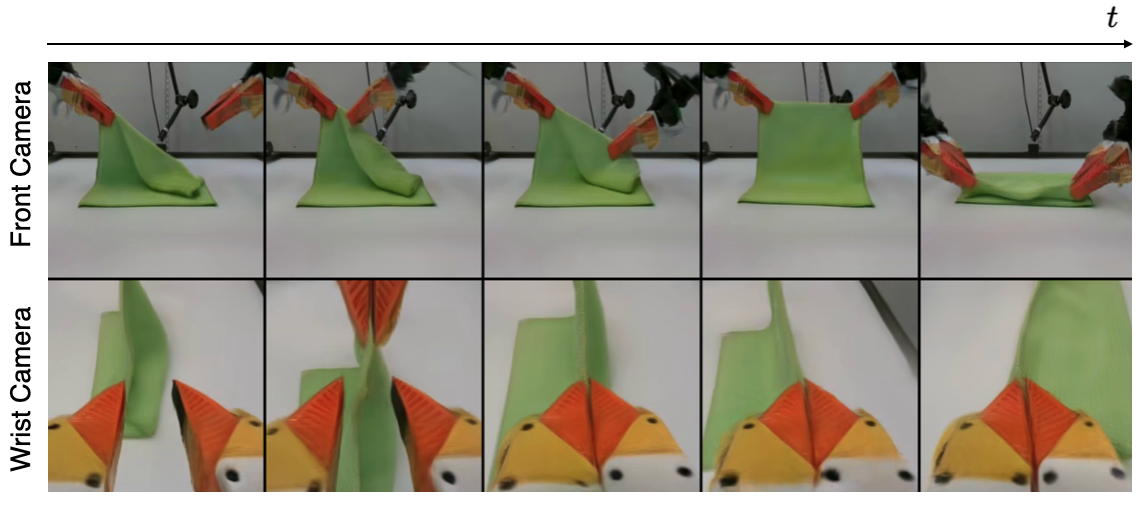}
    \captionof{figure}{\textbf{Visualization of cloth folding world model multi-view input to policies.}}
\end{center}

\begin{center}
    \includegraphics[width=1\linewidth]{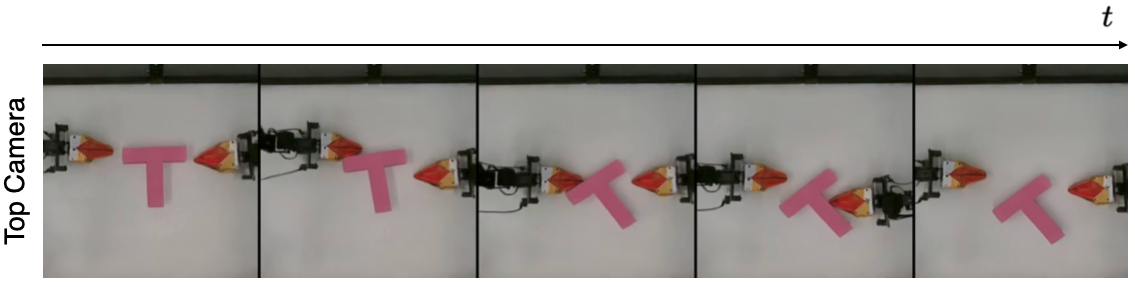}
    \captionof{figure}{\textbf{Visualization of the top down camera view policy input.}}
\end{center}

\end{document}